\documentclass[conference]{IEEEtran}
\IEEEoverridecommandlockouts
\usepackage{cite}
\usepackage{algorithm}
\usepackage{algpseudocode}
\usepackage{amsmath}
\usepackage{amssymb}
\usepackage{graphicx}
\usepackage{textcomp}
\usepackage{tikz}
\usetikzlibrary{positioning, shapes.geometric, arrows.meta, calc, backgrounds, fit}
\usepackage{xcolor}
\usepackage{comment}
\usepackage[pdftex]{hyperref}
\usepackage{titlesec}
\usepackage{graphicx}
\usepackage{booktabs}
\usepackage{caption}
\usepackage{makecell}
\usepackage{subcaption}
\usepackage{ulem}
\usepackage[margin=1in]{geometry}
\usepackage{soul,color}

\def\BibTeX{{\rm B\kern-.05em{\sc i\kern-.025em b}\kern-.08em
    T\kern-.1667em\lower.7ex\hbox{E}\kern-.125emX}}
    
\begin{document}

\title{TimesNet-Gen: Deep Learning-based Site Specific Strong Motion Generation \

}

\author{
\IEEEauthorblockN{1\textsuperscript{st} Baris Yilmaz}
\IEEEauthorblockA{\textit{Multimedia Informatics} \\
\textit{Middle East Technical University}\\
Ankara, Turkey \\
yilmaz.baris\_01@metu.edu.tr}
\and
\IEEEauthorblockN{2\textsuperscript{nd} Bevan Deniz Cilgin}
\IEEEauthorblockA{\textit{Data Informatics} \\
\textit{Middle East Technical University}\\
Ankara, Turkey \\
cilgin.bevan@metu.edu.tr}
\and
\IEEEauthorblockN{3\textsuperscript{rd} Assoc. Prof. Erdem Akagündüz}
\IEEEauthorblockA{\textit{Multimedia Informatics} \\
\textit{Middle East Technical University}\\
Ankara, Turkey \\
akaerdem@metu.edu.tr}
\and
\IEEEauthorblockN{4\textsuperscript{th} Assist. Prof. Salih Tileylioglu}
\IEEEauthorblockA{\textit{Civil Engineering} \\
\textit{Kadir Has University}\\
Istanbul, Turkey \\
salih.tileylioglu@khas.edu.tr}
}

\maketitle

\begin{abstract}
Effective earthquake risk reduction relies on accurate site-specific evaluations, which require models capable of representing the influence of local site conditions on ground motion characteristics. We address strong ground motion generation from time-domain accelerometer records and introduce the TimesNet-Gen, a deep generative framework. In this framework, site-specific generation is directly achieved through a station-restricted, Dirichlet-based latent space resampling strategy, without relying on explicit conditioning inputs or dimensionality reduction. Pre-trained on the AFAD dataset via self-supervised learning, the frozen model demonstrates robust cross-regional generalization by successfully generating station-specific NGA-West2 records without any fine-tuning. Model performance is evaluated by comparing the distributions of generated and real records in the log-HVSR space, alongside the joint analysis of peak ground acceleration and fundamental site frequency. As a baseline, we construct a spectrogram-based conditional variational autoencoder (CVAE) explicitly formulated for station-specific latent space modeling. The results show strong station-wise alignment, consistent cross-regional ground motion synthesis, and a favorable comparison with a spectrogram-based conditional variational autoencoder baseline, demonstrating that the model empirically maintains the essential physical coupling between frequency content and peak amplitude. Our codes are available via \href{https://github.com/brsylmz23/TimesNet-Gen}{a public repo}.
\end{abstract}


\section{Introduction}
Earthquakes have historically led to significant loss of life and extensive economic damage. The 2023 Turkey–Syria earthquakes had a death toll exceeding 53,000, caused more than \$103 billion in economic losses, and affected over 13 million people \cite{undp2025two_years}. Although these impacts cannot be fully eliminated, combining source characterization, recurrence modeling, ground-motion prediction, site-specific hazard assessment, and resilient design practice has substantially reduced earthquake related damage. Among these components, site effects are especially critical because local geology can heavily modify the amplitude, frequency content, and duration of ground shaking \cite{boore1997site}.

Strong motion recordings capture ground acceleration from seismic stations during earthquakes, providing a direct link between underground fracture dynamics and surface effects. While physics-based simulations and empirical ground motion prediction equations (GMPEs) have been widely used to model these motions, they often struggle to fully capture the complex, nonlinear, and site-dependent characteristics inherent in different geographical regions \cite{campbell2014nga}. Data-driven frameworks offer a promising direction by learning directly from data distributions, reducing the need for restrictive parametric assumptions. However, as discussed in \cite{caglar2025exploringchallengesdeeplearning}, there is still no foundational deep learning model that can effectively represent the complex temporal and spectral patterns of these recordings.

In this paper, we address strong ground motion generation from time-domain accelerometer records and introduce TimesNet-Gen, a time-domain generative model built by adapting TimesNet \cite{wu2023timesnettemporal2dvariationmodeling}. We hypothesize that strong motion waveforms can be effectively synthesized in a deep generative framework relying solely on station-restricted latent resampling. While existing studies utilize physical parameters such as magnitude, distance, and velocity \cite{florez2020datadrivenaccelerogramsynthesisusing}, station-restricted sampling isolates site-specific patterns that are characteristically consistent but naturally variable across events.

As a baseline, we train a convolutional VAE \cite{51978} on amplitude/phase spectrograms with station-ID conditioning. In order to benchmark both approaches, site-frequency based analyses are carried out. We utilize strong-motion recordings from the Disaster and Emergency Management Presidency of Türkiye (AFAD) database \cite{turkmen2024deeplearningbasedepicenterlocalization} for both training and initial testing to evaluate the generated site-specific records within this regional context. Furthermore, to explicitly test the cross-regional feature extraction capability of our framework without any regional fine-tuning, we freeze the AFAD-trained network. By projecting NGA-West2 records through this frozen encoder to construct new target latent banks, we conduct additional evaluations using the geographically and tectonically distinct Next Generation Attenuation-West2 (NGA-West2) dataset \cite{bozorgnia2014nga}.

Our contributions are as follows:
\begin{itemize}
    \item[(i)] We introduce TimesNet-Gen, a novel time-domain generative framework that adapts the TimesNet backbone to natively achieve site-specific generation through a station-restricted, Dirichlet-based latent resampling and coupled aggregation strategy.
    \item[(ii)] In addition to training and testing our model on the AFAD dataset, we demonstrate its robust cross-regional generalization by evaluating its transferability on the NGA-West2 dataset.
    \item[(iii)] We evaluate the generative fidelity through distribution-level comparisons in the log-HVSR space and joint $f_{0}$-PGA analysis. This approach verifies that the synthesized records match the empirical spectral envelopes and consistently maintain the physical coupling between frequency content and peak amplitude.
    \item[(iv)] Unlike seismological predictors, our approach learns site-controlled signatures directly from data without extensive parameter tuning or strong theoretical assumptions.
\end{itemize}

\subsection{Related Work}
The related literature on classical and deep learning based seismic data generation and evaluation is summarized below. 

\noindent\textit{1) Classical Methods on Seismic Data Synthesis:} These methods can be broadly categorized into three types: empirical \cite{abrahamson1992stable} \cite{joyner1981peak}, semi-empirical \cite{arora2020strong}, and physics-based \cite{mai2010hybrid}. In general, classical methods involve simplifying assumptions and computational cost for broadband simulations. These approaches often struggle to reproduce the fine-scale variability and nonstationary temporal–spectral behavior of strong-motion recordings, and methods that model such nonstationarity more realistically generally require high computational cost, while faster empirical and semi-empirical approaches capture only limited aspects of it. The drawbacks of the mentioned methods make generative models advantageous for transforming the complex nature of seismic waves into valuable insights, as they learn directly from the data and have a low computational cost. For a general overview of classical and other approaches, the reader may refer to \cite{doi:10.1177/87552930231212475}.\\

\noindent\textit{2) Deep Learning-based Seismic Data Generation:} These generative models provide advanced frameworks for creating synthetic seismic data by learning directly from existing data distributions. In recent years, deep generative models have increasingly been explored for modeling high-dimensional temporal signals, including seismic waveforms and strong ground motions \cite{shen2024aftershock, mengxue2026simulation, ba2025conditional, matsumoto2024site}. By learning directly from data distributions, these approaches reduce the need for restrictive parametric assumptions that are commonly adopted in traditional stochastic ground motion modeling frameworks \cite{graves2010broadband, pousse2006nonstationary}. Variational AutoEncoders (VAEs) have been widely used for this task due to the fact that they can be trained with smaller sets and create a latent space that can be used to easily sample new data \cite{li2020seismic} \cite{rnnvae}. Another deep generative architecture, generative adversarial networks (GANs) can create high-fidelity samples; however, their training is challenging due to instability during training and mode collapse problems \cite{li2020seismicgan} \cite{shi2024broadband} \cite{yamaguchi2024site}. Diffusion models generate data from random noise by iteratively learning the reverse of a noise perturbation process. However, they require high computational power due to the numerous refinement steps involved \cite{bergmeister2024highresolutionseismicwaveform} \cite{jung2025broadbandgroundmotionsynthesis}. Although these developments demonstrate promising capabilities and improved flexibility compared to conventional approaches, current generative frameworks still face challenges in consistently reproducing the full nonlinear and site-dependent complexity of earthquake ground motions. Therefore, despite notable recent advances, a broadly accepted and physically comprehensive generative foundation model for strong ground motion simulation has not yet been established.\\

\noindent\textit{3) Evaluation:} Evaluating the output of generative models is a complex task. The primary objective of a deep generative model is typically to implicitly or explicitly model the training set distribution. However, assessing individual signals in terms of realism, diversity, or other qualities requires careful analysis. While off-the-shelf methods exist for vision-based generation, such as Fréchet Inception Distance (FID) \cite{heusel2017gans}, or text-based generation, including BLEU and ROUGE scores \cite{lin2004rouge}, there are no established standards for evaluating generated seismic data due to the limited number of studies in this domain.

Regarding the evaluation of simulated seismic data \cite{chen2017stochastic}, prior works commonly assess intensity measures, frequency-domain similarity, or performance improvements in downstream tasks. Common methods for evaluating generated seismic samples include comparing the intensity measures (IMs) of generated records with those of real records \cite{yamaguchi2024site} \cite{bergmeister2024highresolutionseismicwaveform}, assessing metrics in the frequency domain and waveform shape \cite{matsumoto2024generative} \cite{jung2025broadbandgroundmotionsynthesis}, and measuring the impact of generated samples on the performance of main tasks, such as waveform classification \cite{li2020seismicgan}.

However, existing approaches often struggle to fully verify whether generative models preserve the physical relationships inherent in real earthquake events. In this work, we validate our generative framework through distribution-level assessments in the log-HVSR space and a joint analysis of peak ground acceleration and fundamental site frequency ($f_0$). This dual approach ensures that the generated ground motions preserve both the broad spectral envelope of the target site and the essential physical dependency between spectral content and amplitude.

\begin{figure*}[t]
\centering
\includegraphics[width=\textwidth]{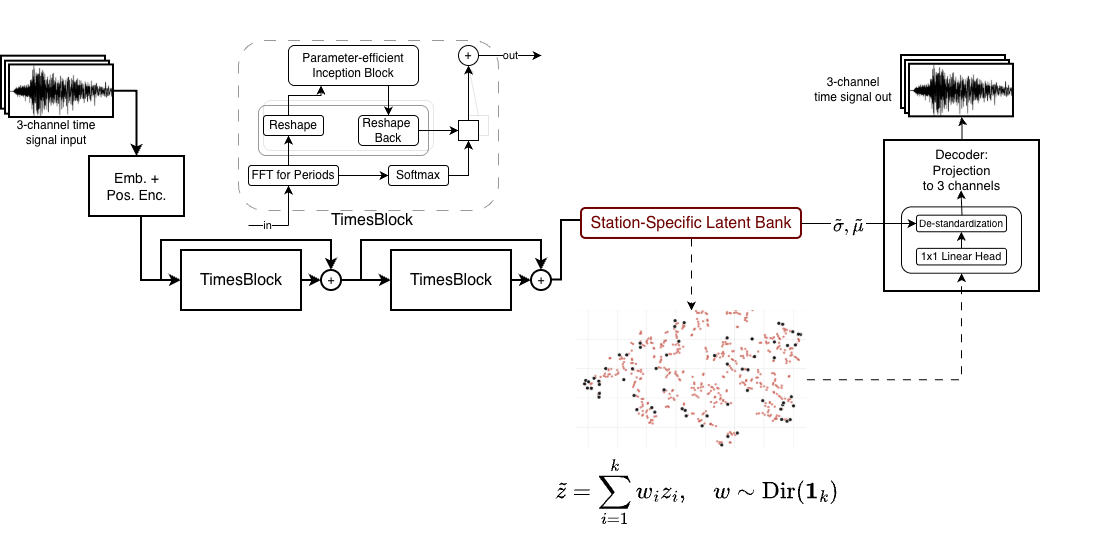}
\caption{The proposed TimesNet-Gen architecture.}
\label{fig:timesnetgen_architecture}
\end{figure*}

\section{Methodology}

We benchmark two approaches for strong-motion reconstruction and conditional generation: the proposed TimesNet-Gen and a conditional VAE baseline. Below, the details of the architectures, the Dirichlet-based sampling method for station conditioning, input data preprocessing steps, and the utilized datasets are presented. 

\subsection{TimesNet-Gen}

The original TimesNet model \cite{wu2023timesnettemporal2dvariationmodeling} introduces temporal 2D-variation modeling for general time series, where the 1D sequence is decomposed into multiple period-aligned 2D slices by selecting top-$k$ dominant periods from the frequency domain. For each selected period $p$, the sequence is reshaped into a $(T/p, p)$ grid so that intraperiod-variation (within a period) appears along columns and interperiod-variation (across periods at the same phase) appears along rows. Each grid is processed by a parameter-efficient, Inception-style 2D convolutional branch that captures temporal patterns across short and long contexts. Branch outputs are aggregated with soft, period-dependent weights and merged through a residual path; a lightweight linear layer then projects features back to the original number of signal components (e.g., three acceleration channels). When clear periodicity is absent, variations are dominated by intra-period structure, which the backbone handles as the limiting case of very long periods. Prior work used this backbone for forecasting, imputation, classification, and anomaly detection.

Unlike generic time-series models that focus solely on these discriminative tasks, strong ground motion synthesis requires a framework capable of preserving site-specific spectral signatures while allowing for physically bounded stochastic variability. We selected TimesNet as our foundational backbone because its unique ability to explicitly model periodic 2D structures aligns perfectly with the mixed-wave nature (P-waves, S-waves, and surface waves) of seismic records. Building on this backbone, we develop a time-domain conditional generative model, namely ``TimesNet-Gen'', for reconstruction and station-specific generation. To transform this architecture into a generative foundation, the adaptation preserves the original encoder, incorporates a linear decoder head for signal reconstruction, and utilizes a high-dimensional latent representation where station-specific characteristics are exclusively modeled via the latent sampling space  (Figure \ref{fig:timesnetgen_architecture}). During the self-supervised pre-training phase, the network learns to reconstruct acceleration waveforms directly from the time series using a Mean Squared Error (MSE) reconstruction loss, $\mathcal{L} = \|x - \hat{x}\|^2$, while preserving critical broadband spectral properties.

The backbone follows the standard TimesNet temporal block design, applying FFT-based period selection and multi-kernel 2D convolutions. We operate directly on multi-channel acceleration records $x \in \mathbb{R}^{T \times C}$ (N–S, E–W, V) at 100 Hz. Sequences are center-aligned: if longer than the required dimension, a center crop around the midpoint is taken; if shorter, symmetric zero-padding is applied around the midpoint to reach the target sequence length. Finally, we apply per-sequence, per-channel standardization and inverse-transform the outputs before computing time-domain losses to ensure the original physical scale is strongly preserved.

Before selecting this architecture, we considered other transformer-based models \cite{nie2022time} and multi-scale architectures \cite{wang2024timemixer}. However, these methods primarily learn temporal dependencies without explicitly considering periodic structures. TimesNet-Gen, conversely, reshapes the one-dimensional time series into a two-dimensional representation based on dominant periods, explicitly capturing intra-period and inter-period variations. This aligns perfectly with the characteristics of seismic signals, which consist of mixed wave components (P-waves, S-waves, surface waves) with distinct frequency ranges.

To construct a robust generative framework, we extract the encoder output tensor from the TimesNet backbone and achieve station-restricted generation through a Dirichlet-based latent space resampling formulation. This mechanism enforces site-specific representation while ensuring sufficient sampling diversity to generate physically realistic and highly variable strong ground motions.

\subsubsection{Conditioning and Dirichlet-based Latent Space Sampling}
Unlike conventional conditional generative models that utilize explicit class embeddings (e.g., one-hot encoded vectors) at the encoder level, TimesNet-Gen achieves site-specific generation natively by restricting latent draws to the target station’s bank. The framework therefore relies exclusively on a Dirichlet-based latent resampling strategy to enforce station identity and naturally control generation diversity.

Latent space sampling is a method of creating new signal representations by choosing and combining representative latent vectors. By controlling the weighting and summation of latent components, this process directly influences the variability of the generated synthetic ground motions. To generate diverse and physically realistic strong ground motions, the model extracts representations from real records in a self-supervised manner to construct a station-specific latent bank. 

During the generation phase for a given station, a subset of $k$ latent vectors $(z_1, \dots, z_k)$ is randomly drawn with bootstrap sampling from the corresponding station's latent pool. In our experiments, the number of selected latent vectors is defined as $k = \min(5, N)$, where $N$ indicates the total number of available latent vectors for that target station.

From this stage onward, weights are drawn from a uniform symmetric Dirichlet distribution on the $(k-1)$-simplex (equivalently, $\mathrm{Dir}(\mathbf{1}_k)$ on the $k$-dimensional probability simplex), such that their sum equals one:

\begin{equation}
(w_1, \dots, w_k) \sim \mathrm{Dir}(\mathbf{1}_k), \quad \sum_{i=1}^{k} w_i = 1.
\label{eq:dirichlet}
\end{equation}

The aggregated latent representation is then obtained through a convex combination:
\begin{equation}
\tilde{z} = \sum_{i=1}^{k} w_i z_i
\end{equation}

This latent vector, $\tilde{z}$, captures the site-specific spectral characteristics. To map the generated signals back to their physical amplitude scales, we utilize the sample-specific mean ($\mu_{i} \in \mathbb{R}^{3}$) and standard deviation ($\sigma_{i} \in \mathbb{R}^{3}$) vectors. These parameters, originally calculated per-record and per-channel across time as a reversible instance normalization step, are aggregated from the latent bank using the same Dirichlet weights:

\begin{equation}
\tilde{\mu} = \sum_{i=1}^{k} w_i \mu_i, \quad \tilde{\sigma} = \sum_{i=1}^{k} w_i \sigma_i
\label{eq:mixed_stats}
\end{equation}

A time-independent linear projection processes $\tilde{z}$ to output a normalized waveform. The final three-channel acceleration signal is then obtained by rescaling per channel $(c \in \{N-S, E-W, V\})$ through an inverse standardization step:

\begin{equation}
x_{\text{gen}}^{(c)}[t] = \big(\mathrm{Linear}(\tilde{z}[t])\big)^{(c)} \cdot \tilde{\sigma}^{(c)} + \tilde{\mu}^{(c)}
\label{eq:inverse_standardization}
\end{equation}

Applying identical weights to both the latent features and their scaling parameters maintains the physical dependency between spectral content and amplitude. Relying on a global variance parameter would decouple these properties, mitigating the risk of generating non-physical records. Our analyses indicate that this coupled aggregation effectively guides the generated ground motions to preserve the inherent aleatory variability and amplitude bounds of the target station.

\subsection{Variational Autoencoders}
Variational Autoencoders (VAEs) \cite{kingma2022autoencodingvariationalbayes} are generative models that learn a probabilistic mapping between observed data and a lower-dimensional latent space. They consist of an encoder and a decoder that compress and reconstruct the data, enabling the generation of new samples. The proposed variational autoencoder consists of a convolutional encoder and a symmetric deconvolutional decoder. The encoder employs four consecutive convolutional layers with 3x3 kernels and Leaky ReLU activations, followed by a flattening operation and two linear layers that produce the latent mean and standard deviation parameters. Latent variables are sampled using the reparameterization trick and passed to the decoder, which begins with a linear transformation and reshaping step. The decoder comprises four transposed convolutional layers with 3x3 kernels and Leaky ReLU activations, concluding with a sigmoid output layer that reconstructs the input.\\
   
\noindent\textit{1) Latent Space Conditioning:} While TimesNet-Gen conditions through Dirichlet sampling, the baseline VAE is conditioned using a two-phase training approach. In the first sub-phase, the model is conditioned using one-hot encoded class priors. Each class is assigned a distinct prior mean vector $\mu_p$ constructed by tiling one-hot encodings across the latent dimensions, while the prior variance is fixed to $\sigma_p^2 = 1$. To ensure separable priors, the latent dimension is chosen as a multiple of the number of classes (five in this case), resulting in a 510-dimensional latent space. The Kullback--Leibler divergence term used in the loss is computed as

\begin{equation}
\begin{aligned}
\mathrm{KL}(q(z \mid c)\,\|\,p(z \mid c))
= \frac{1}{2} \sum \Big(
&\frac{\sigma_q^2}{\sigma_p^2}
+ \frac{(\mu_q - \mu_p)^2}{\sigma_p^2} \\
&- 1
- \log \frac{\sigma_q^2}{\sigma_p^2}
\Big)
\end{aligned}
\label{eq:one_hot_kl}
\end{equation}

\noindent where $\mu_q$ and $\sigma_q^2$ denote the encoder-predicted mean and variance, respectively, and $(\mu_p, \sigma_p^2)$ correspond to the parameters of the class-specific prior distribution. Here, $p(z \mid c)$ represents the class-conditioned prior. 

In the second phase, the latent space is further refined to sharpen the class clusters, following the approach of Mousavi et al.~\cite{mousavi}. Encoded representations $z_i$ are obtained from the first phase, and initial cluster centers $\mu_j$ are computed as the mean of samples belonging to each class. Cluster membership probabilities are estimated as

\begin{equation}
q_{ij} = \frac{(1 + \| z_i - \mu_j \|^2)^{-1}}{\sum_j (1 + \| z_i - \mu_j \|^2)^{-1}},
\label{eq:membership}
\end{equation}

and a sharpened target distribution is defined as

\begin{equation}
p_{ij} = \frac{\frac{q_{ij}^2}{\sum_i q_{ij}}}{\sum_j \left( \frac{q_{ij}^2}{\sum_i q_{ij}} \right)}.
\label{eq:target_prob}
\end{equation}

The clustering loss is given by the Kullback--Leibler divergence between the sharpened and soft assignment distributions,

\begin{equation}
\mathcal{L}_c = \mathrm{KL}(P \| Q) = \sum_i \sum_j p_{ij} \log \left( \frac{p_{ij}}{q_{ij}} \right),
\label{eq:post_kl}
\end{equation}

and is added to the original VAE objective with a weighting factor $a$ (100 in our experiments), yielding the final loss function
\begin{equation}
\mathcal{L}_{\text{final}} = \mathcal{L}_{\text{VAE}} + a \, \mathcal{L}_c.
\end{equation}\\

\noindent\textit{2) Input Preprocessing:} The aim of these preprocessing steps is to ensure that both amplitude and phase information can be effectively learned by the model while preserving the essential structural characteristics of the signal. In the preprocessing pipeline, spectrograms containing both amplitude and phase information were generated for each seismic record. First, short-time Fourier transform (STFT) is applied to all three channels of each record to produce amplitude and phase spectrograms, resulting in a multi-channel representation for each recording. The amplitude spectrograms are then converted to a logarithmic decibel (dB) scale to make them more perceptually meaningful and to facilitate the learning process. Phase spectrograms, initially obtained as wrapped phases within the range of $-\pi$ to $\pi$ due to the FFT, exhibited discontinuities along the time axis; these are corrected using phase unwrapping to produce a continuous and smooth phase profile.

In order to be able to evaluate the results and compare them to TimesNet-Gen outputs, the spectrograms are converted back to time-domain signals. This was achieved by first denormalizing the amplitude and phase spectrograms, transforming the amplitude spectrograms from the logarithmic scale back to linear values, and then combining the amplitude and phase components before applying the inverse short-time Fourier transform.

\subsection{Datasets}
To evaluate both site-specific generation quality and cross-regional transferability, we employ two distinct strong-motion databases.

\subsubsection{AFAD Dataset}
For self-supervised training and primary evaluation, we utilize strong-motion recordings from the Disaster and Emergency Management Presidency of Türkiye (AFAD) database \cite{turkmen2024deeplearningbasedepicenterlocalization} (36,417 records, 2012–2018). To analyze generative fidelity, we chose five stations with different site properties and fundamental site frequencies. Some of the fundamental frequencies reported here (specifically for stations 2020 and 0205) differ from what is reported in the AFAD website due to differences in the calculation method. Fundamental site frequency is typically calculated empirically from the horizontal-to-vertical spectral ratio (HVSR) analysis of ambient recordings \cite{nakamura1989method} as well as earthquake recordings \cite{lermo1993site} \cite{yazdi2022new}. In this study, we compute it based on strong motions signals \cite{lermo1993site} \cite{yazdi2022new}. The five AFAD stations used in this study and their identified fundamental frequencies are given in Table \ref{tab:ft_stations}. 

\begin{table}[h]
\centering
\caption{Properties of the Selected AFAD Stations}
\begin{tabular}{lccc}
\hline
\textbf{Station Id} & \textbf{Location}     & \textbf{$f_0$} & \textbf{No. of rec.} \\ 
\hline
2020 & Tavas / Denizli       & 5.1 Hz  & 71  \\
4628 & Afsin / Kahramanmaras & 1.8 Hz  & 38  \\
0205 & Kahta / Adiyaman      & 2.6 Hz  & 98  \\
1716 & Ayvacik / Canakkale   & 6.4 Hz  & 110 \\
3130 & Defne / Hatay         & 12.8 Hz & 31  \\
\hline
\end{tabular}
\label{tab:ft_stations}
\end{table}

\subsubsection{NGA-West2 Dataset (Cross-Regional Evaluation)}
To evaluate the cross-regional generalization of TimesNet-Gen, we test its performance on a strong ground motion dataset collected from a tectonically and geographically distinct region. For this purpose, we use a subset of the NGA-West2 database for Southern California provided by the Pacific Earthquake Engineering Research Center (PEER) \cite{bozorgnia2014nga}.

To ensure compatibility with future physical simulator evaluations, we selected only records where complete metadata is available, including fault mechanism parameters such as strike and rake, which are required inputs for the SCEC Broadband Platform simulator \cite{maechling2015scec}. The resulting subset contains approximately 8,000 ground motion records from 893 stations. For evaluation, we prioritized diversity in site conditions ($V_{s30}$ values) and ensured sufficient record counts per station. The selected NGA-West2 stations are detailed in Table \ref{tab:nga_stations}.

\begin{table}[h]
\centering
\caption{Properties of the Selected NGA-West2 Stations}
\begin{tabular}{lccc}
\hline
\textbf{Station Name} & \textbf{$V_{s30}$ (m/s)} & \textbf{$f_0$ (Hz)} & \textbf{No. of rec.} \\
\hline
Barrett & 511 & 4.9 & 66 \\
Palomar & 465 & 6.2 & 66 \\
Chilao Flat Rngr & 927 & 8.3 & 70 \\
Domenigoni Reservoir & 576 & 2.1 & 67 \\
Mt. San Jacinto Campus & 244 & 1.8 & 43 \\
\hline
\end{tabular}
\label{tab:nga_stations}
\end{table}

\subsection{Evaluation}
To comprehensively analyze the generation quality and physical realism of the synthesized strong ground motions, we employ a multi-tiered evaluation framework comprising spectral analysis and distribution-level metrics.

\subsubsection{Spectral Analysis}
The characteristics of earthquake recordings depend heavily on site effects, which modify and amplify ground motions through resonance \cite{kramer2024geotechnical}. To evaluate this, we compute the Horizontal-to-Vertical Spectral Ratio (HVSR) curves using signals recorded from the three components: north-south ($H_{NS}(t)$), east-west ($H_{EW}(t)$), and vertical ($V(t)$). First, the amplitude spectra $|H_{NS}(f)|$, $|H_{EW}(f)|$, and $|V(f)|$ are calculated. The average horizontal amplitude is defined as:
\begin{equation}
|H(f)| = \sqrt{ \frac{ |H_{NS}(f)|^2 + |H_{EW}(f)|^2 }{2} }
\label{eq:hvsr_horiz}
\end{equation}
The HVSR curve is then constructed by dividing the average horizontal amplitude by the vertical amplitude spectra for each frequency:
\begin{equation}
\text{HVSR}(f) = \frac{ |H(f)| }{ |V(f)| }
\label{eq:hvsr_ratio}
\end{equation}
From this curve, the fundamental site frequency is identified as the peak frequency, $f_0 = \arg\max_f \text{HVSR}(f)$. We restrict our analysis to the 1--20 Hz band and apply the same preprocessing and smoothing across all sources to ensure consistency. Additionally, the horizontal Peak Ground Acceleration (PGA) is extracted from the time domain as the maximum magnitude of the combined horizontal acceleration vectors over the entire record length, defined as $\text{PGA}_h = \max_t \left( \sqrt{a_{EW}(t)^2 + a_{NS}(t)^2} \right)$ to capture amplitude consistency.

\subsubsection{Interstation Discrimination}
To evaluate the overall physical consistency of the generated signals, we initially analyze the statistical distributions of the calculated $f_0$ values obtained from real, reconstructed, and model-generated samples for each station. To quantify the similarity between these distributions, we employ the Jensen-Shannon Divergence ($D_{JS}$), converted into a similarity score where 1 indicates perfect similarity and 0 indicates none. These scores are compiled into a confusion-style matrix to evaluate inter-station discrimination. To assess the overall quality of this clustering, we compute the distance between our resulting matrices and an ideal identity-block matrix using Normalized Cross-Correlation (NCC).

\subsubsection{Physical Consistency}
To ensure the generative process adheres to the physical data manifold, we evaluate the structural agreement between the synthesized and empirical ground motions. This is achieved through two complementary consistency checks.

First, we assess the broadband spectral envelope by comparing the distributions of log-HVSR values. For each station, the horizontal-to-vertical ratios are mapped to a logarithmic scale, and a histogram-based density estimate is constructed to represent the probability distribution. We then quantify the similarity between the generated and empirical distributions. Let $p$ and $q$ denote the normalized histogram probability vectors of $\log$-HVSR values for the two distributions $P$ and $Q$, with a small uniform smoothing mass $\varepsilon$ applied before renormalization to ensure numerical stability. We compute the Jensen-Shannon distance using a base-2 logarithm, denoted as $d_{\mathrm{JS}}(P,Q) \in [0,1]$. This quantity is a proper metric and satisfies the triangle inequality. We define the reported similarity score as:
\begin{equation}
S_{\mathrm{JS}}(P,Q) = 1 - d_{\mathrm{JS}}(P,Q).
\label{eq:js-sim}
\end{equation}

High $S_{JS}$ scores indicate that the model effectively reproduces the full spectral envelope characteristics of the target site.

Second, we qualitatively assess the physical coupling between spectral content and peak amplitude by visualizing the bivariate joint distribution of $f_{0}$ and horizontal PGA. In empirical seismic events, these parameters are strongly coupled. By observing the overlap between the generated and real records in the $f_{0}$--PGA feature space, we confirm that the synthesized outputs maintain the essential physical dependency inherent in the training distribution.

\section{Experiments and Results}

To comprehensively evaluate the proposed framework, the experimental results are structured into two main phases. First, we conduct an evaluation using the training database (AFAD) to assess latent space characteristics, spectral consistency, and interstation discrimination. Subsequently, we perform a cross-regional generalization test using the NGA-West2 database to evaluate the model's transferability to geographically distinct regions.

\subsection{Intra-Regional Evaluation (AFAD Dataset)}

\subsubsection{Latent Space Representation}

\begin{figure}[h]
\centering
\includegraphics[width=0.9\columnwidth,clip]{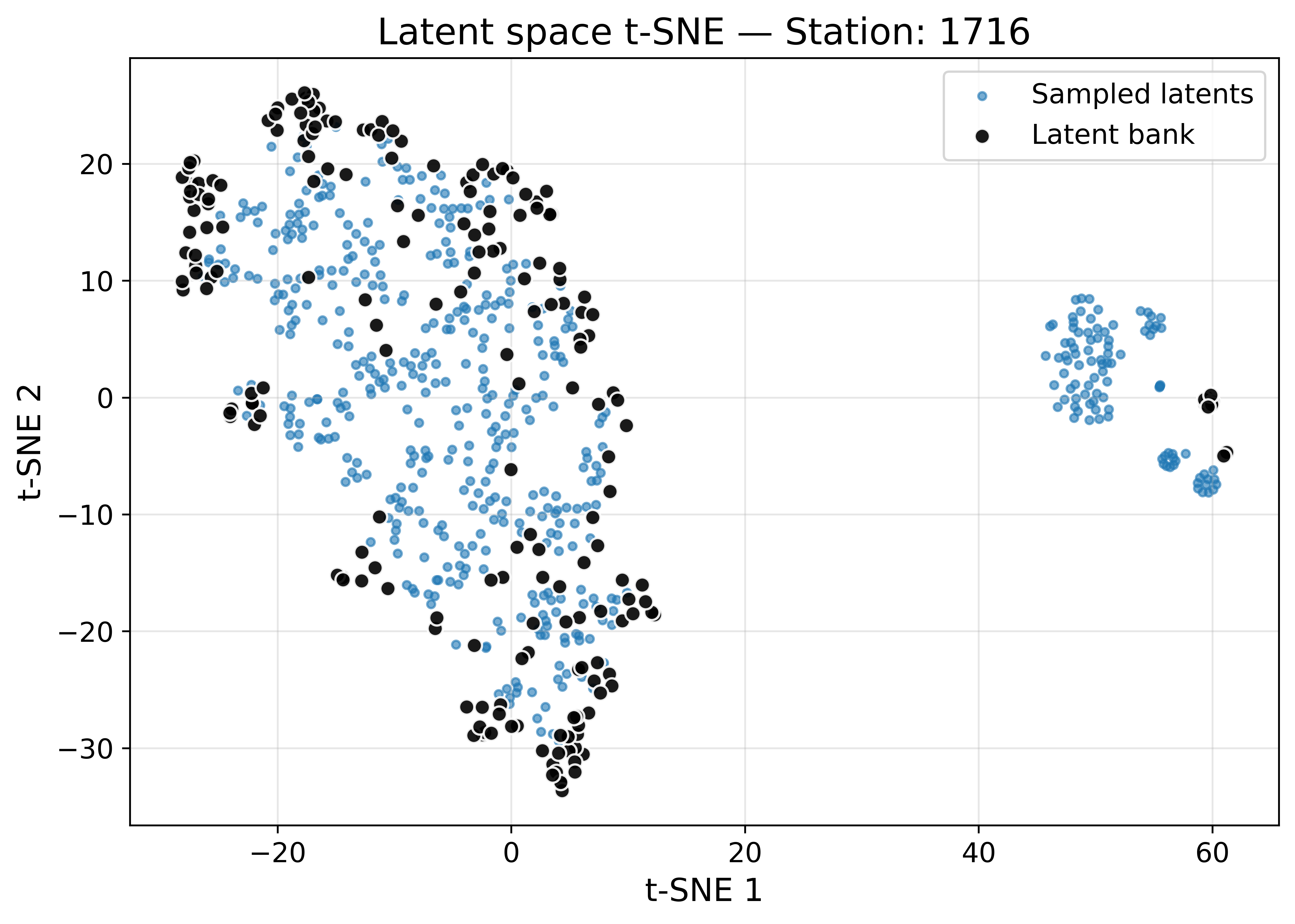}
\caption{t-SNE projection of the latent space for station 1716.}
\label{fig:latent_tsne_uniform_1716}
\end{figure}

Figure \ref{fig:latent_tsne_uniform_1716} illustrates the $t$-SNE projection of the latent representations for station 1716 to evaluate the coherence of the learned seismic features. Black points correspond to the latent bank derived from real records, while blue points represent generated samples. While t-SNE is an illustrative projection and not strictly metric-preserving, this visualization reveals a qualitatively continuous distribution of latent features. This suggests that the latent space provides a diverse manifold for seismic variations, where generated samples effectively cover the domain of empirical records without forming isolated, non-physical artifacts.

\begin{figure}[!htbp]
\centering
\includegraphics[width=\columnwidth]{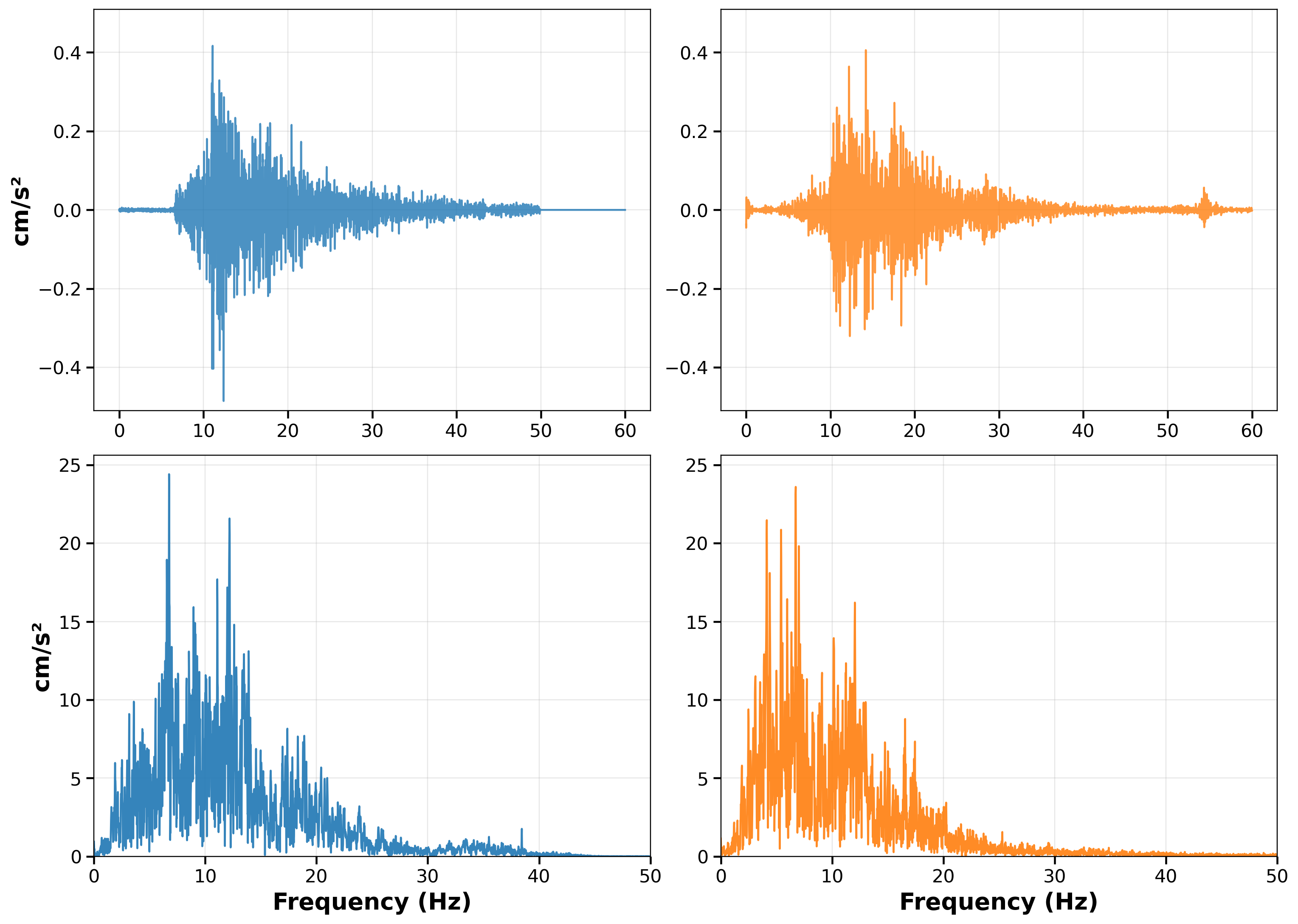}
\vspace{2mm}
\includegraphics[width=\columnwidth]{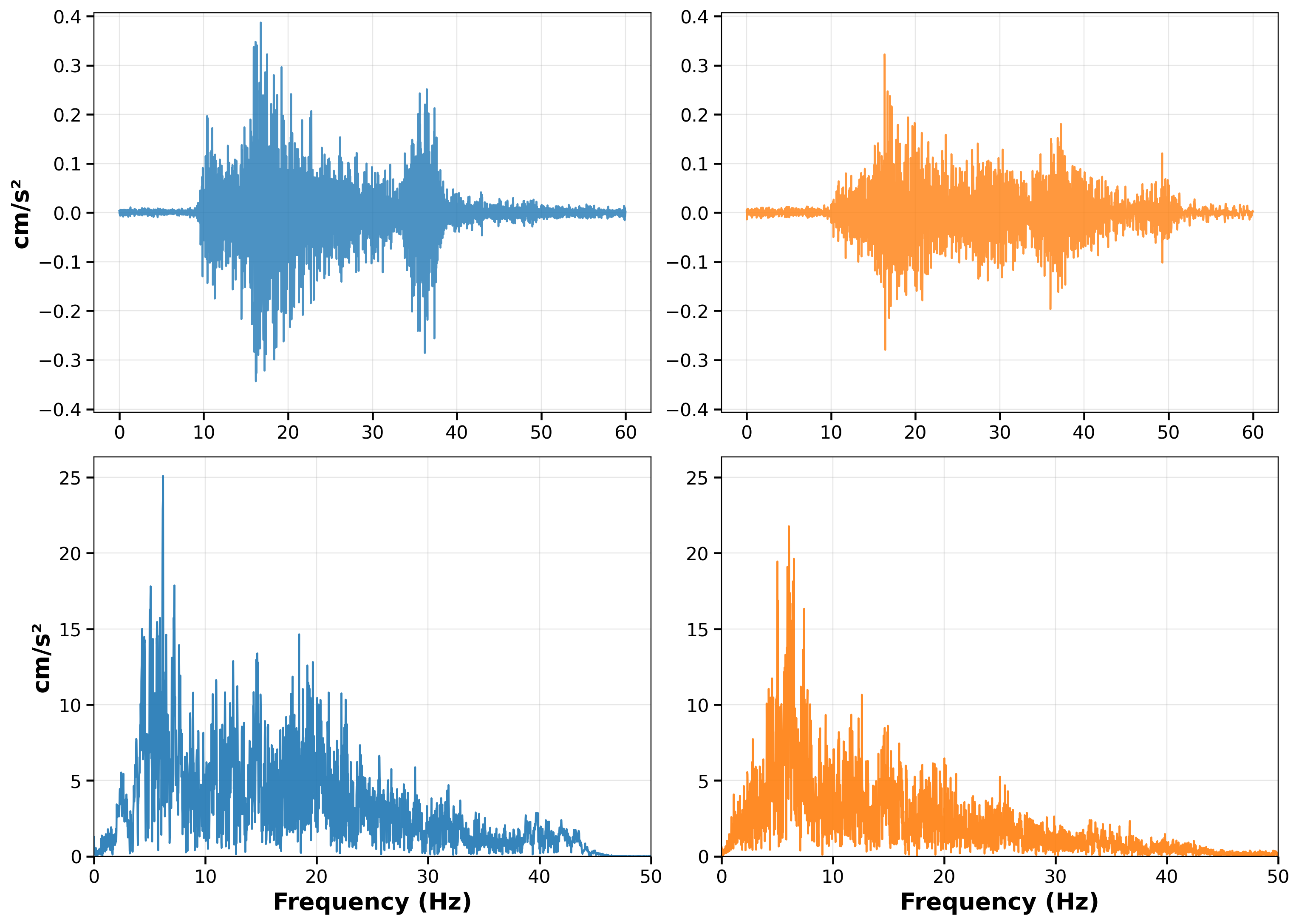}
\vspace{2mm}
\includegraphics[width=\columnwidth]{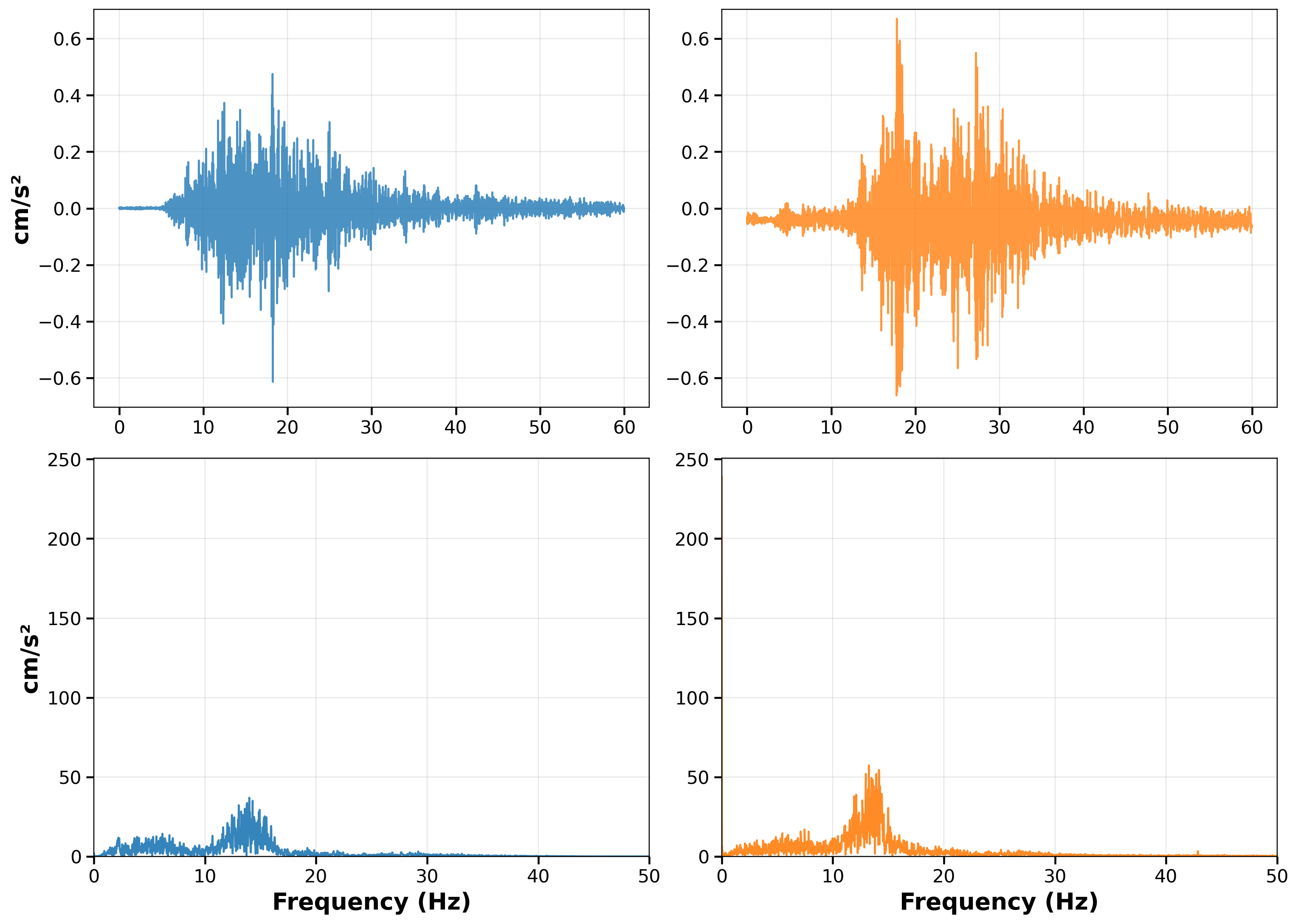}
\caption{Real and TimesNet-Gen generated samples for three stations with corresponding Fourier amplitude spectra.}
\label{fig:ts_examples}
\end{figure}

To assess the temporal and spectral characteristics of the synthesized signals, Figure \ref{fig:ts_examples} compares real and TimesNet-Gen generated E-W components. The generated waveforms exhibit realistic durations and amplitudes, with Fourier amplitude spectra consistent with the empirical data.

\subsubsection{Spectral Consistency and Site Response}

\begin{figure*}[!p]
\centering

\begin{subfigure}{0.38\textwidth} 
    \centering
    \includegraphics[width=\linewidth]{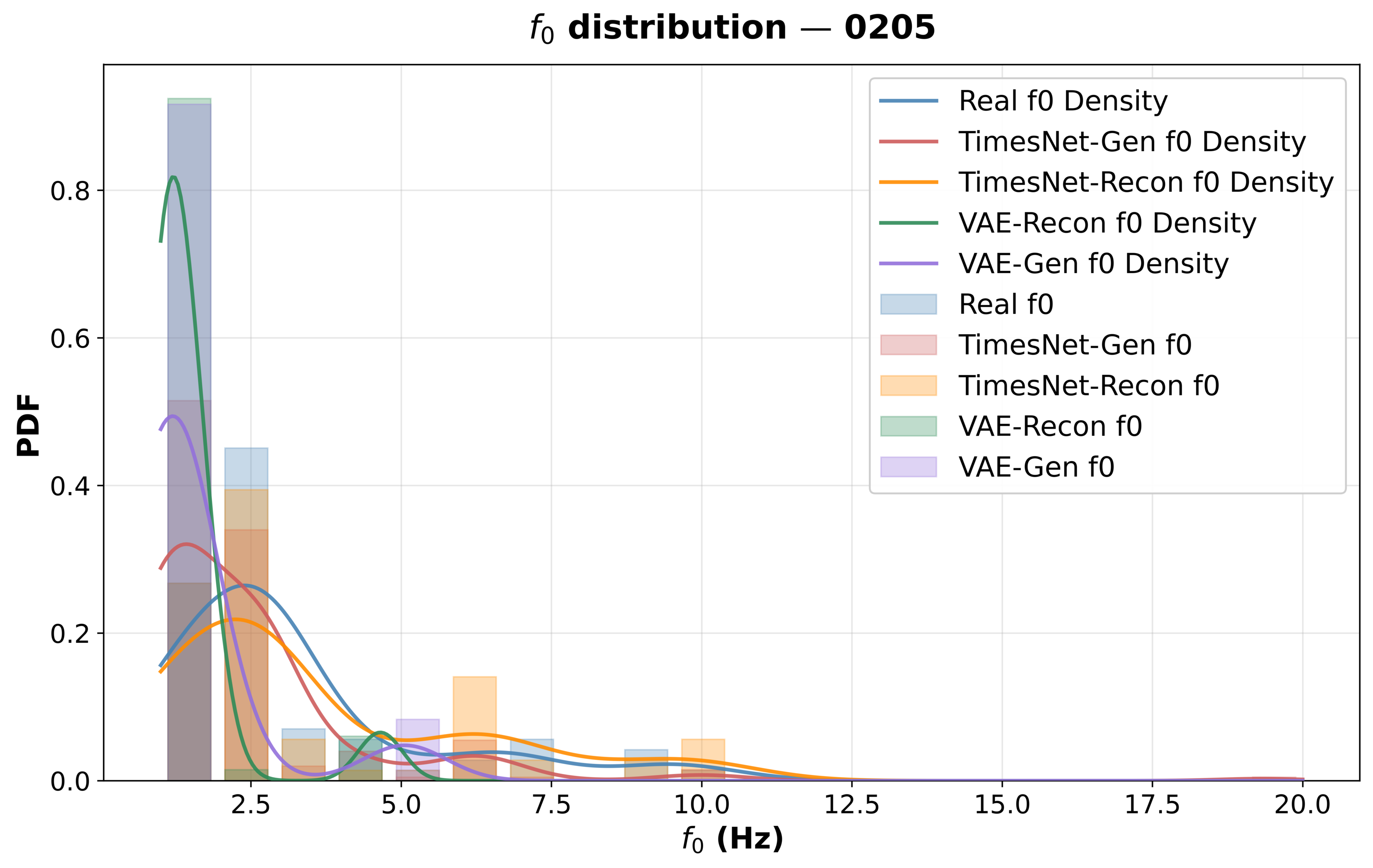}
    \caption{Station 0205 - \texorpdfstring{$f_0$}{f0}}
\end{subfigure}
\hspace{0.5cm} 
\begin{subfigure}{0.42\textwidth}
    \centering
    \includegraphics[width=\linewidth]{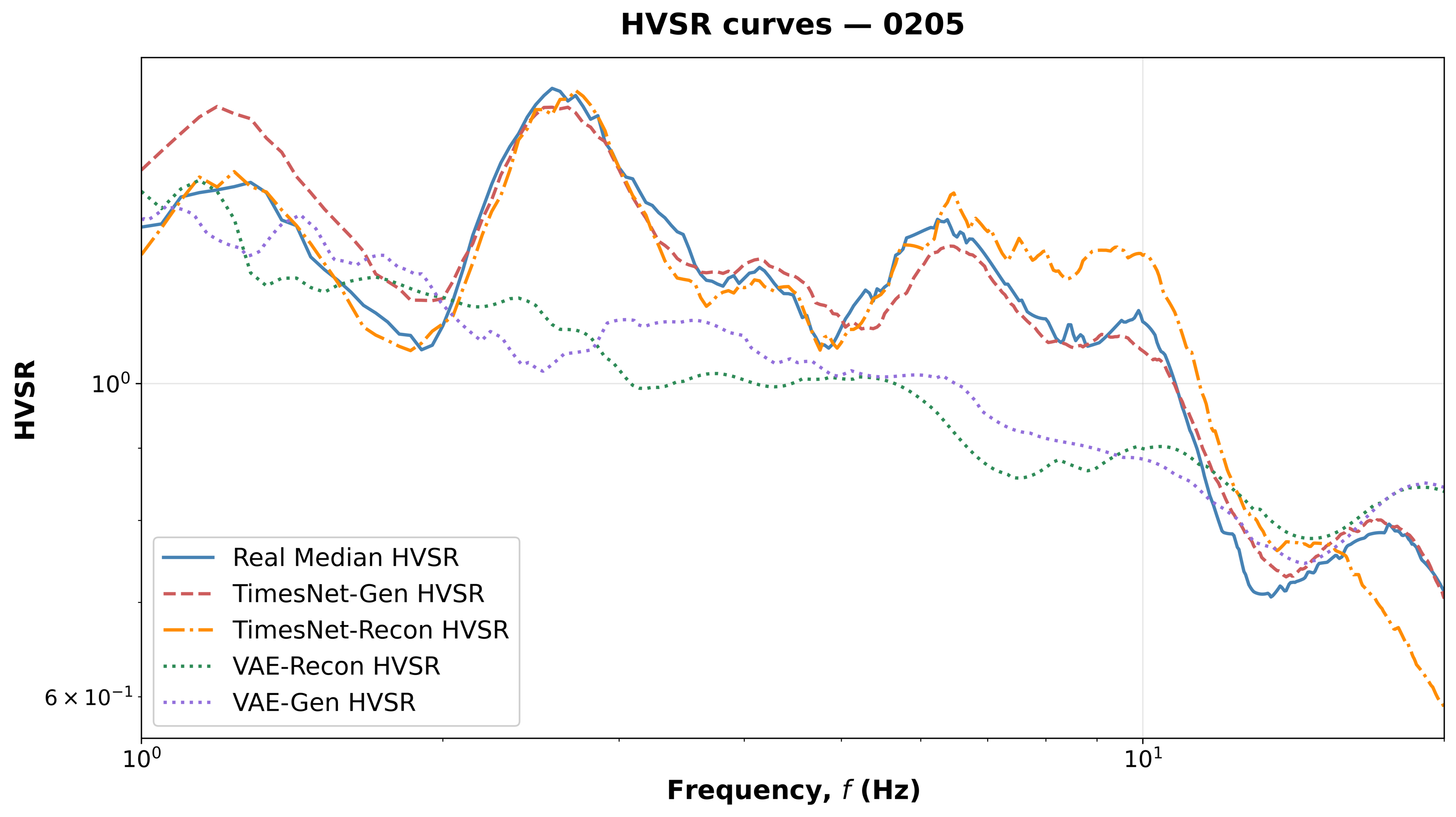}
    \caption{Station 0205 - HVSR}
\end{subfigure}

\vspace{0.15cm} 

\begin{subfigure}{0.38\textwidth}
    \centering
    \includegraphics[width=\linewidth]{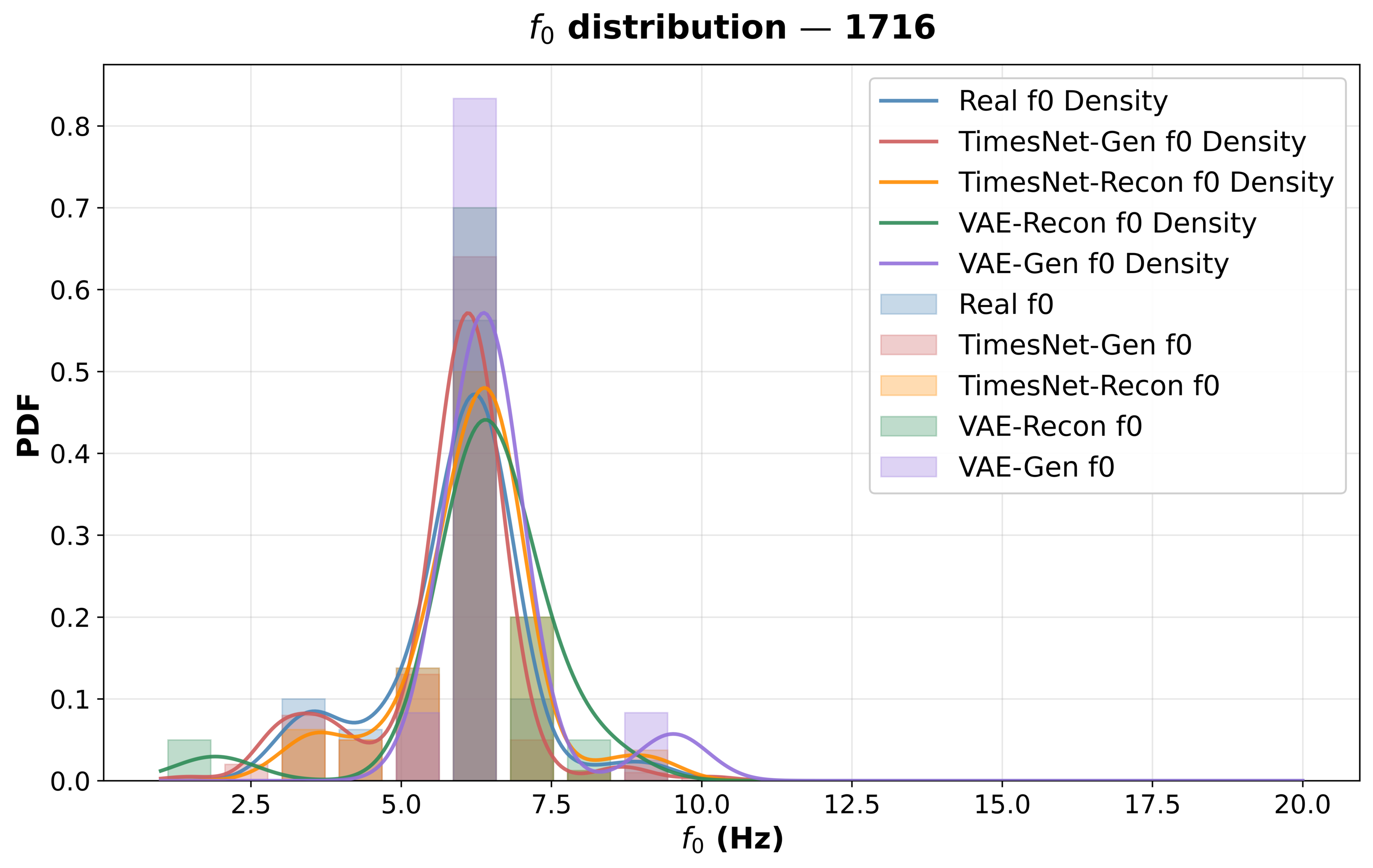}
    \caption{Station 1716 - \texorpdfstring{$f_0$}{f0}}
\end{subfigure}
\hspace{0.5cm}
\begin{subfigure}{0.42\textwidth}
    \centering
    \includegraphics[width=\linewidth]{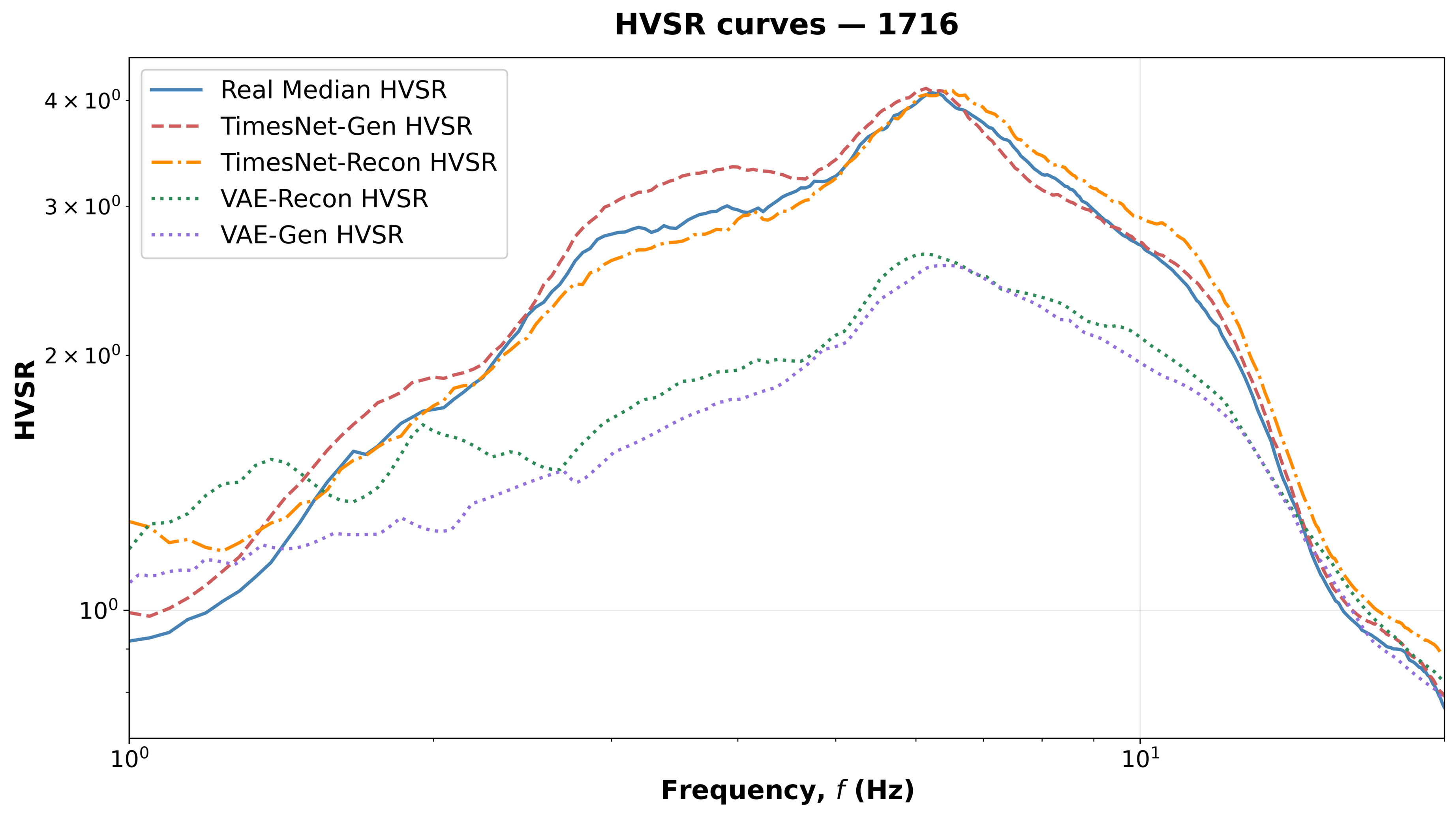}
    \caption{Station 1716 - HVSR}
\end{subfigure}

\vspace{0.15cm}

\begin{subfigure}{0.38\textwidth}
    \centering
    \includegraphics[width=\linewidth]{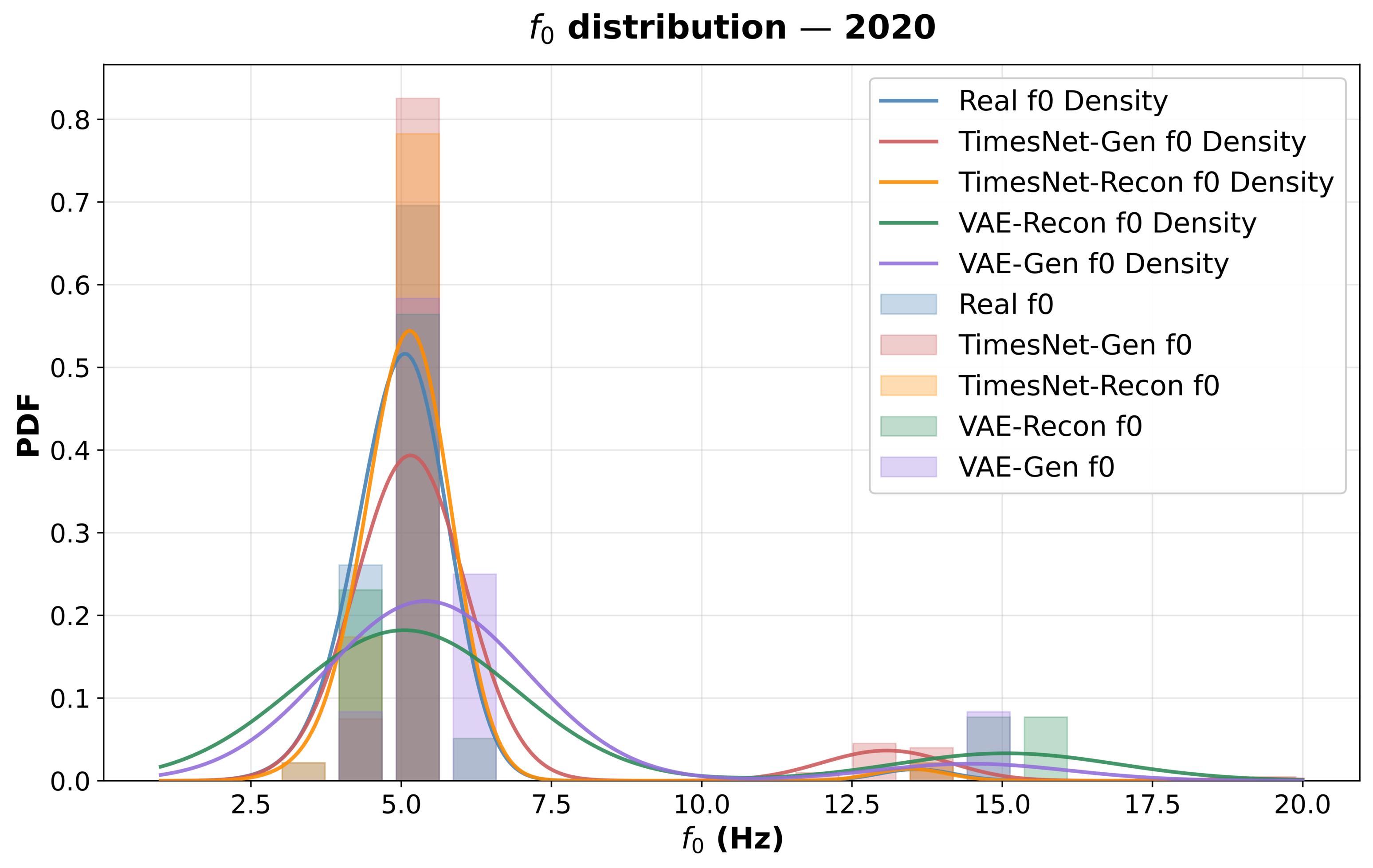}
    \caption{Station 2020 - \texorpdfstring{$f_0$}{f0}}
\end{subfigure}
\hspace{0.5cm}
\begin{subfigure}{0.42\textwidth}
    \centering
    \includegraphics[width=\linewidth]{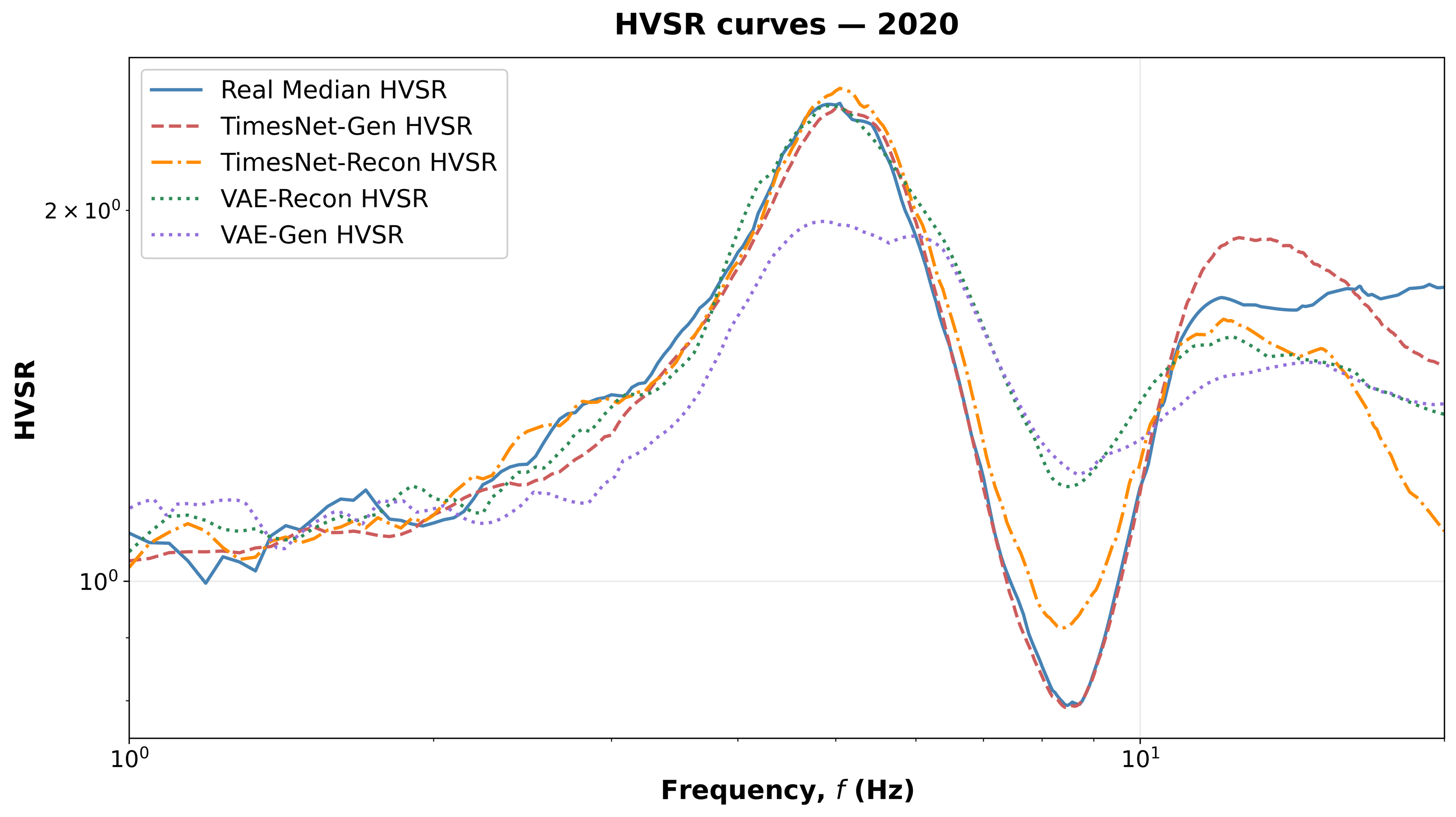}
    \caption{Station 2020 - HVSR}
\end{subfigure}

\vspace{0.15cm}

\begin{subfigure}{0.38\textwidth}
    \centering
    \includegraphics[width=\linewidth]{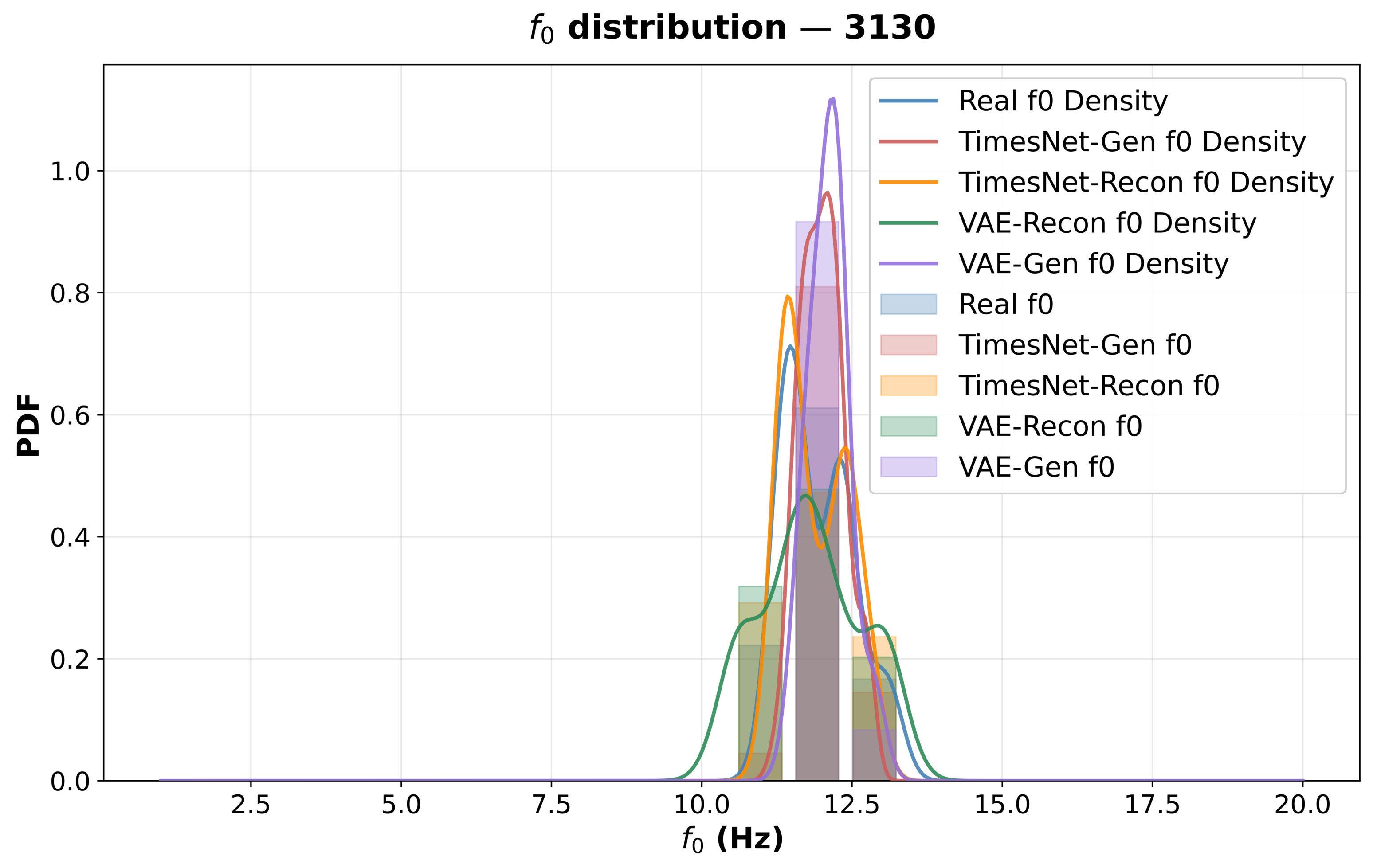}
    \caption{Station 3130 - \texorpdfstring{$f_0$}{f0}}
\end{subfigure}
\hspace{0.5cm}
\begin{subfigure}{0.42\textwidth}
    \centering
    \includegraphics[width=\linewidth]{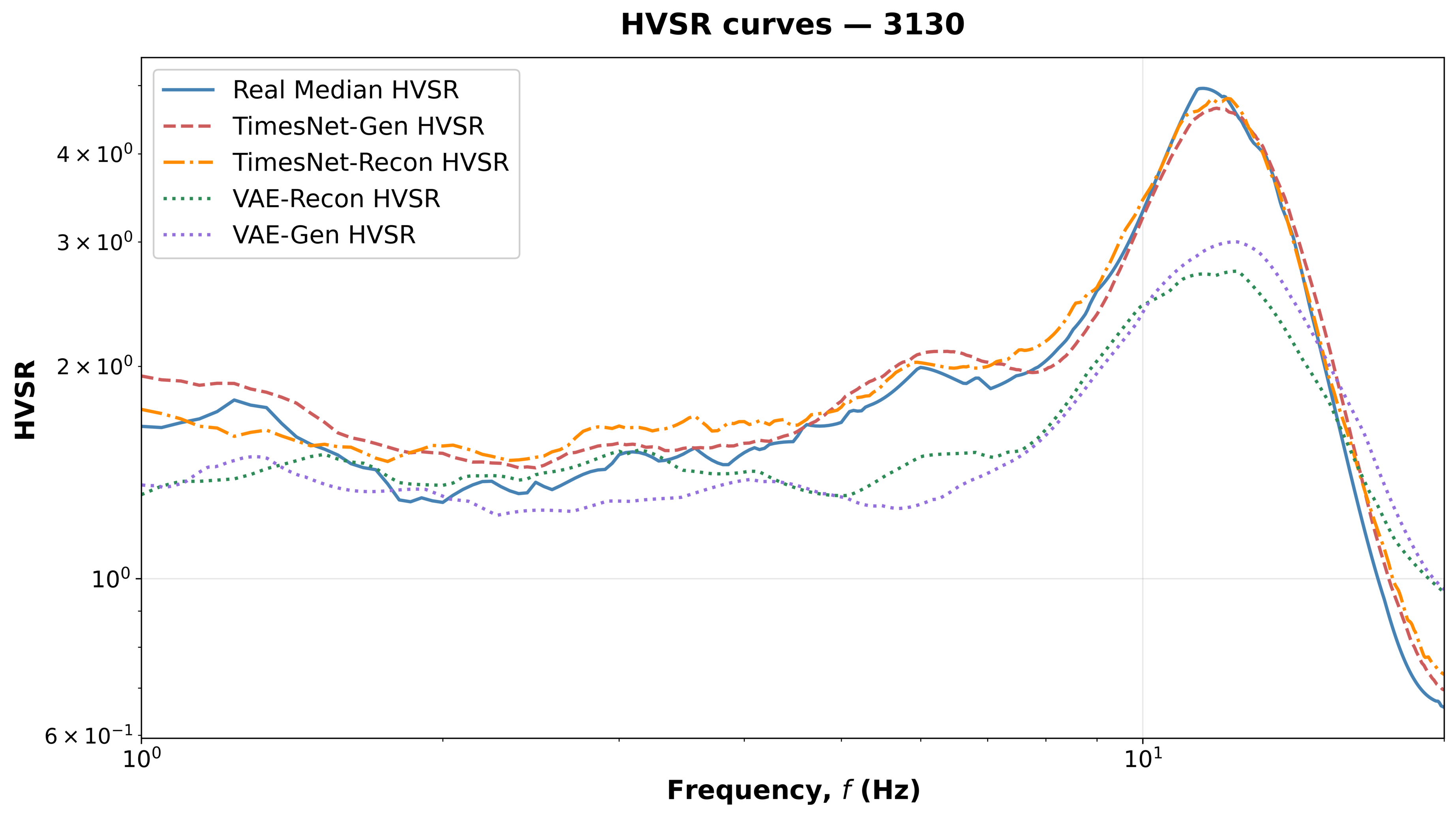}
    \caption{Station 3130 - HVSR}
\end{subfigure}

\vspace{0.15cm}

\begin{subfigure}{0.38\textwidth}
    \centering
    \includegraphics[width=\linewidth]{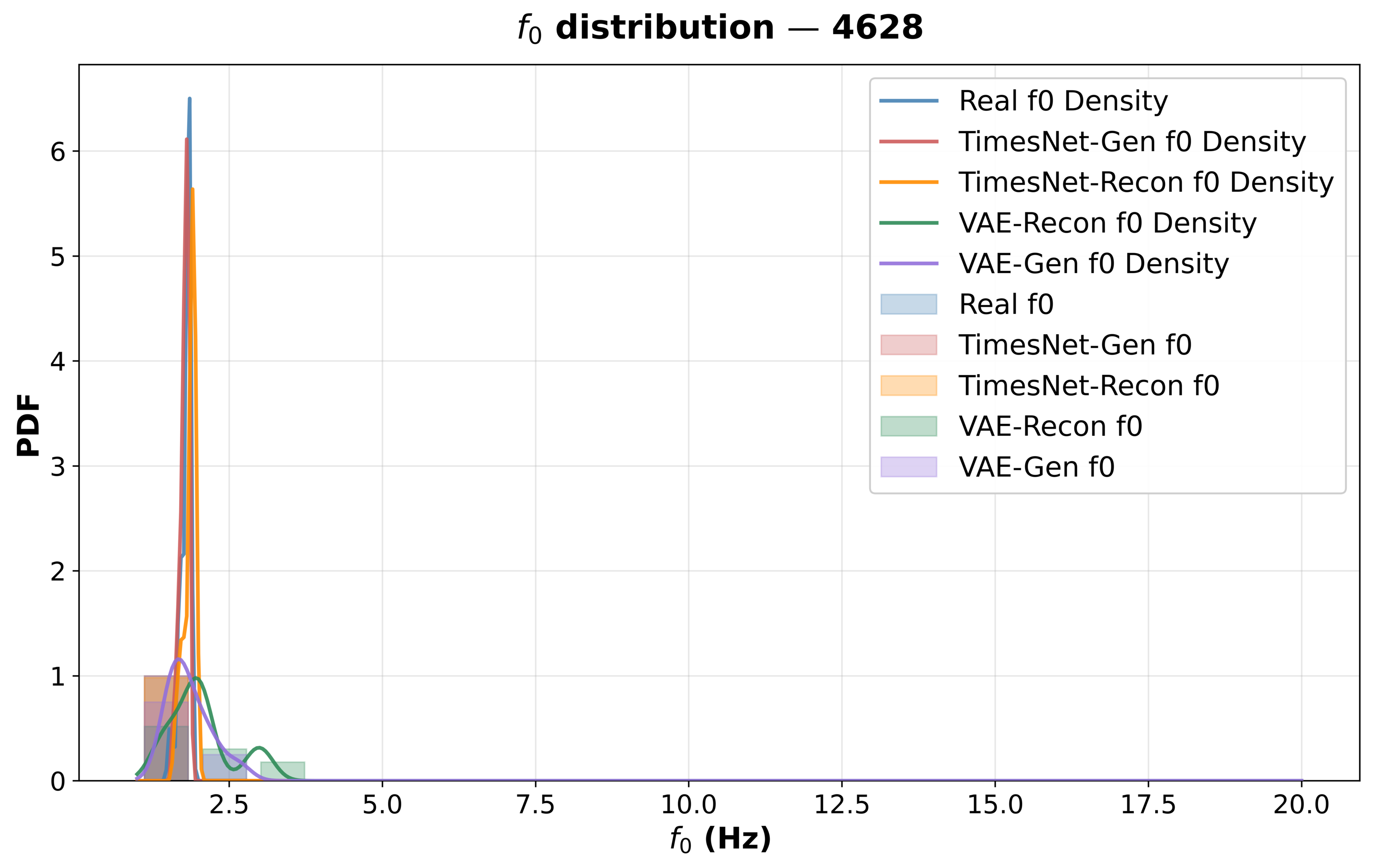}
    \caption{Station 4628 - \texorpdfstring{$f_0$}{f0}}
\end{subfigure}
\hspace{0.5cm}
\begin{subfigure}{0.42\textwidth}
    \centering
    \includegraphics[width=\linewidth]{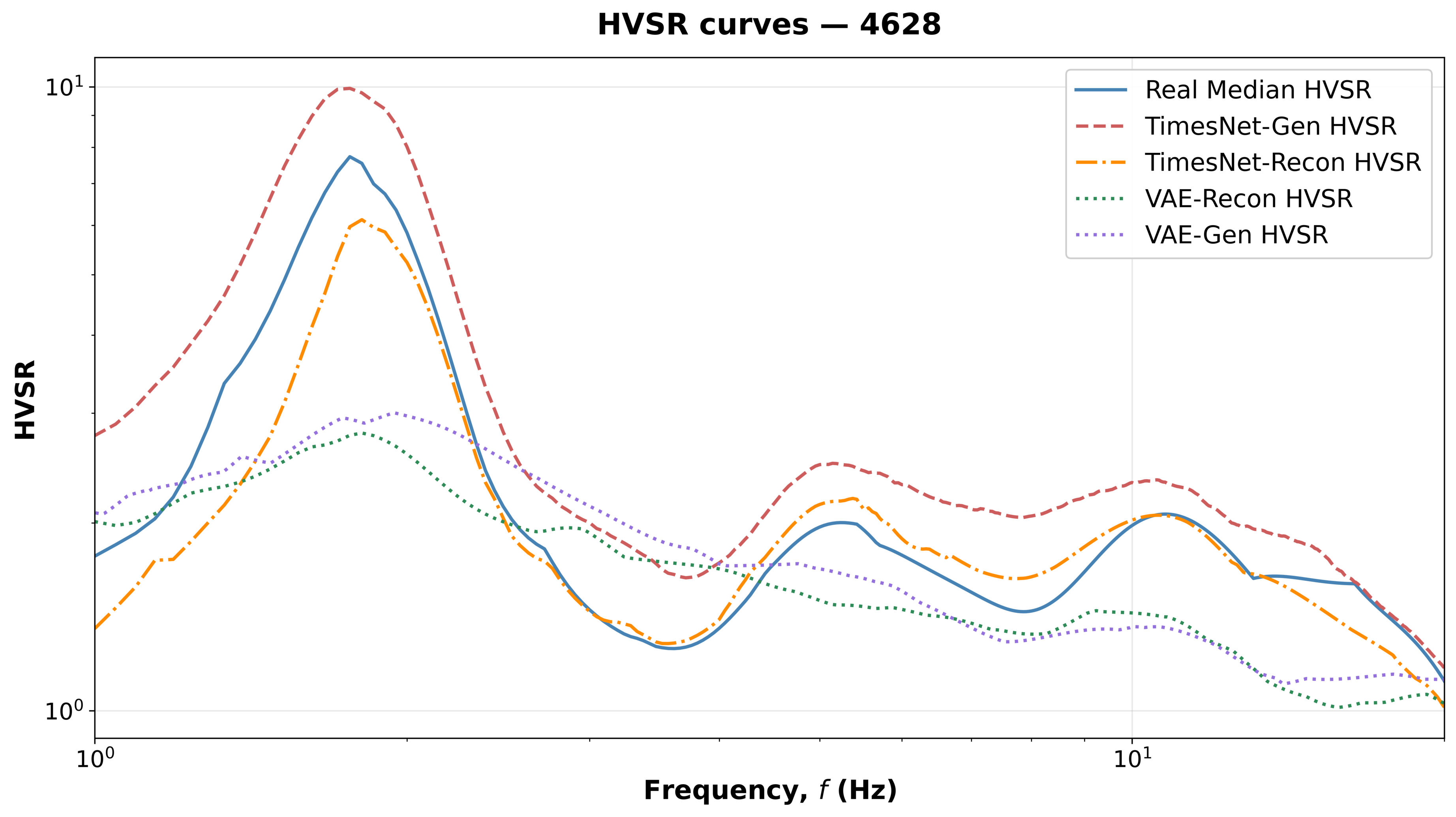}
    \caption{Station 4628 - HVSR}
\end{subfigure}

\caption{Comparison of \texorpdfstring{$f_0$}{f0} distributions (left) and average HVSR curves (right) for the selected AFAD stations.}
\label{fig:combined_spectral_analysis}
\end{figure*}

To assess spectral consistency, we compare the $f_0$ distributions of real data, generated samples, and reconstructed signals (Figure \ref{fig:combined_spectral_analysis}, left column). For each station, 50 synthetic records were generated to construct representational empirical distributions. Reconstructed signals processed through the encoder-decoder pathway establish the performance ceiling for the generative sampling. As shown in the $f_0$ distributions of Figure \ref{fig:combined_spectral_analysis}, TimesNet-Gen demonstrates consistent reconstruction quality across stations and generates samples that closely follow the real $f_0$ distributions, even for sites exhibiting irregular wide-spread behavior. Conversely, the VAE baseline shows considerable performance degradation.

Furthermore, Figure \ref{fig:combined_spectral_analysis} presents the average HVSR curves. TimesNet-Gen captures the distinct peak behaviors, including sharp resonance characteristics at stations 0205 and 4628. In contrast, the VAE deviates significantly from the actual peaks and struggles to reproduce sharp spectral features. The empirical matching of these stable peaks with the target station's $f_{0}$ distribution indicates that the generator reliably recovers the fundamental site-response characteristics.

\subsubsection{Interstation Discrimination}

\begin{figure}[!htbp] 
    \centering
    \captionsetup[subfigure]{justification=centering}

    \begin{subfigure}{0.9\linewidth}
        \centering
        \includegraphics[width=0.8\linewidth, trim=0 9 35 12, clip]{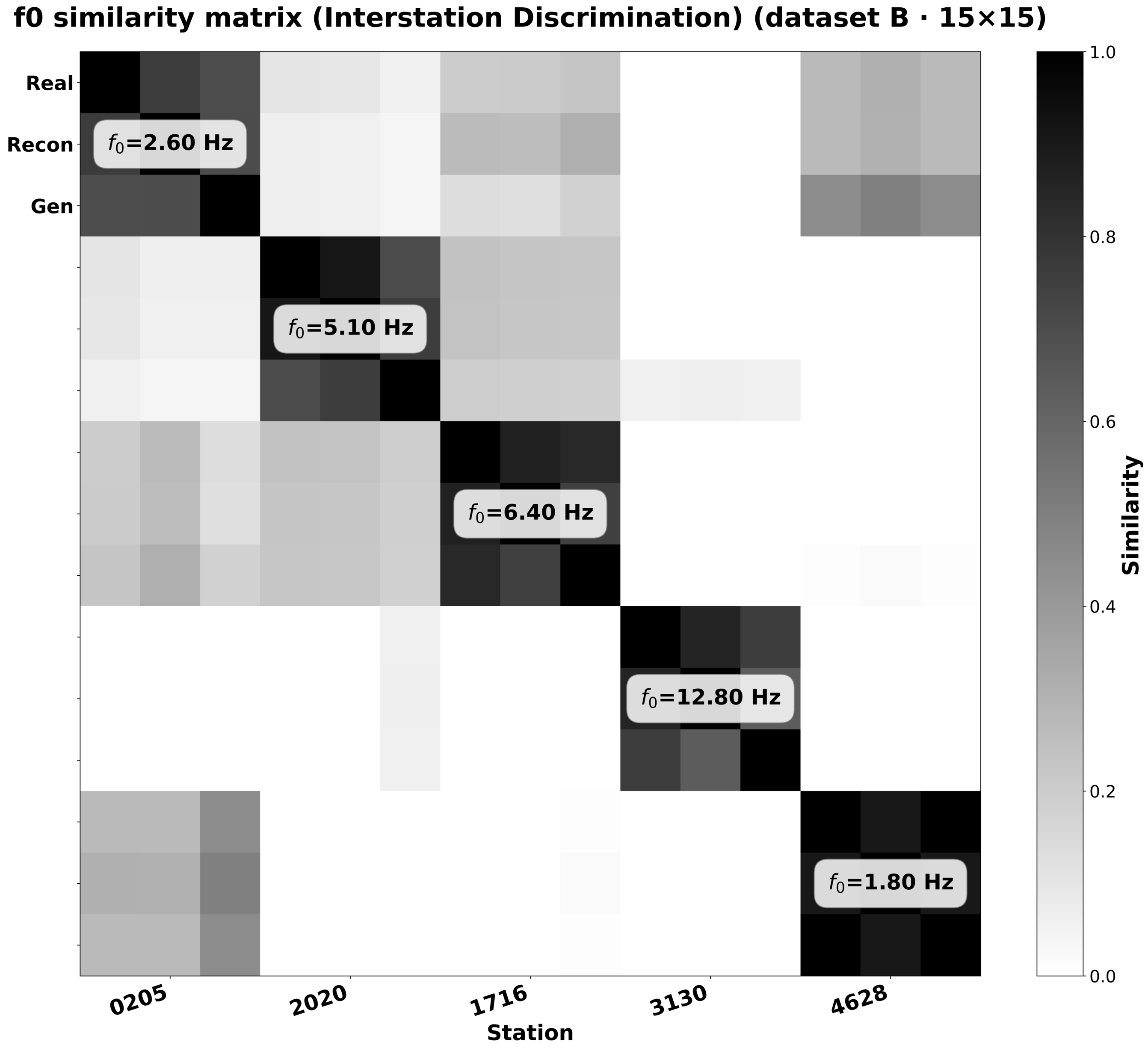}
        \caption{TimesNet-Gen (alignment score: 0.93)}
        \label{fig:timesnet_extended}
    \end{subfigure}
    
    \vspace{0.4cm}
    
    \begin{subfigure}{0.9\linewidth}
        \centering
        \includegraphics[width=0.8\linewidth, trim=0 9 35 12, clip]{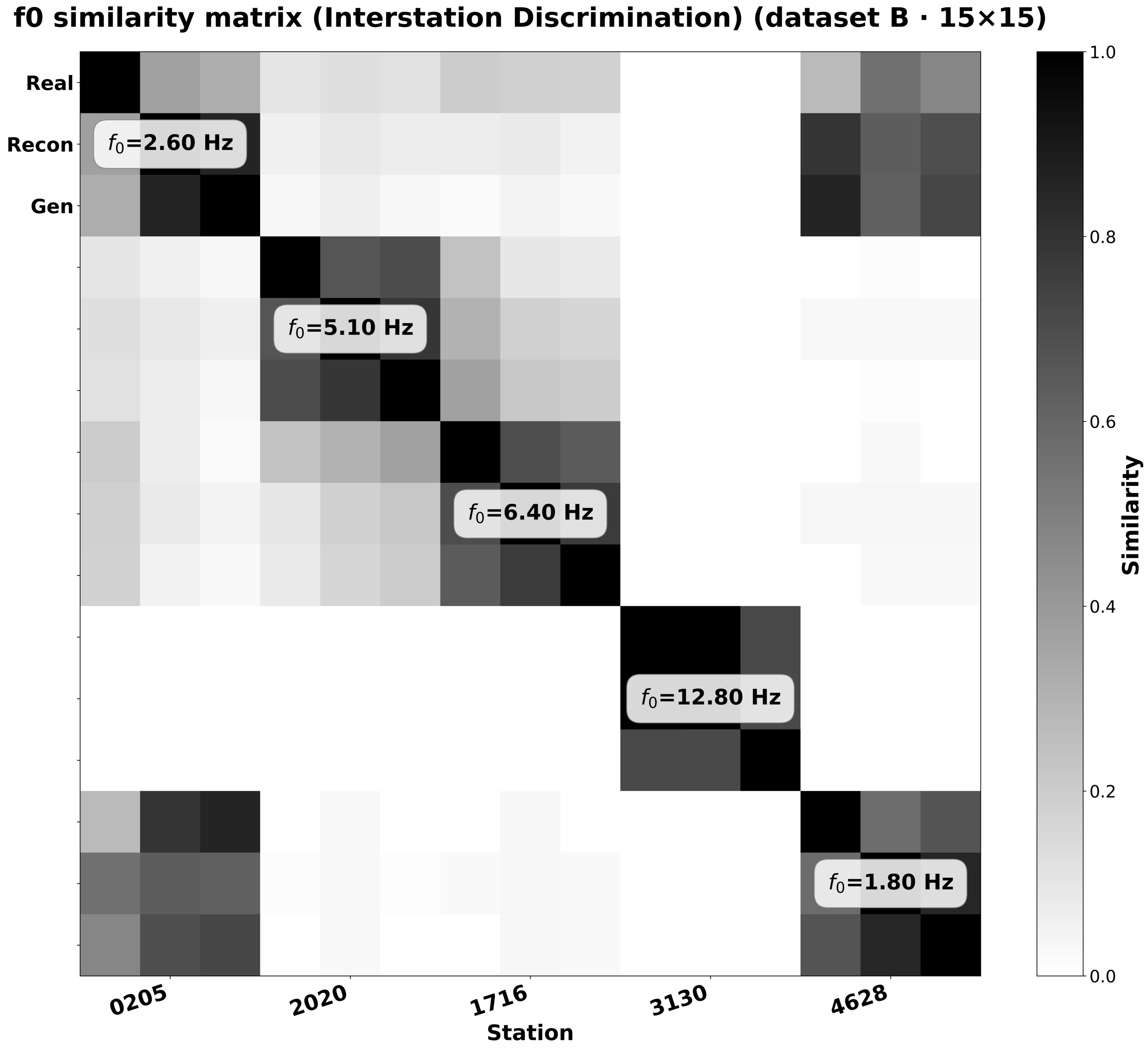}
        \caption{VAE (alignment score: 0.81)}
        \label{fig:vae_extended}
    \end{subfigure}

    \caption{\texorpdfstring{$f_0$}{f0} distribution confusion matrices for (a) TimesNet-Gen and (b) VAE models.}
    \label{fig:extended_js_comparison}
\end{figure}

To evaluate model similarity and interstation discrimination, $f_0$ distribution confusion matrices are analyzed (Figure \ref{fig:extended_js_comparison}). Several stations, notably 0205/4628 and 1716/2020, share comparable $f_0$ values (Table \ref{tab:ft_stations}). While similar $f_0$ values reflect comparable dominant resonance frequencies, the overall site responses can significantly differ. TimesNet-Gen successfully distinguishes these stations despite their $f_0$ value, indicating that it captures complex waveform characteristics beyond the fundamental resonance. The VAE baseline exhibits weaker separation and tends to mix stations that share similar $f_0$ values.

\subsection{Cross-Regional Evaluation (NGA-West2 Dataset)}

To evaluate the cross-regional transferability of TimesNet-Gen, additional experiments are conducted on the NGA-West2 dataset. Specifically, we utilize the frozen AFAD-pretrained encoder to construct new, station-specific latent banks from empirical Southern California records. This protocol assesses the model's generalized feature extraction capability on a geographically distinct region, confirming transferability without relying on regional weight updates.

\subsubsection{Latent Space Transferability}

\begin{figure}[h]
\centering
\includegraphics[width=0.9\columnwidth]{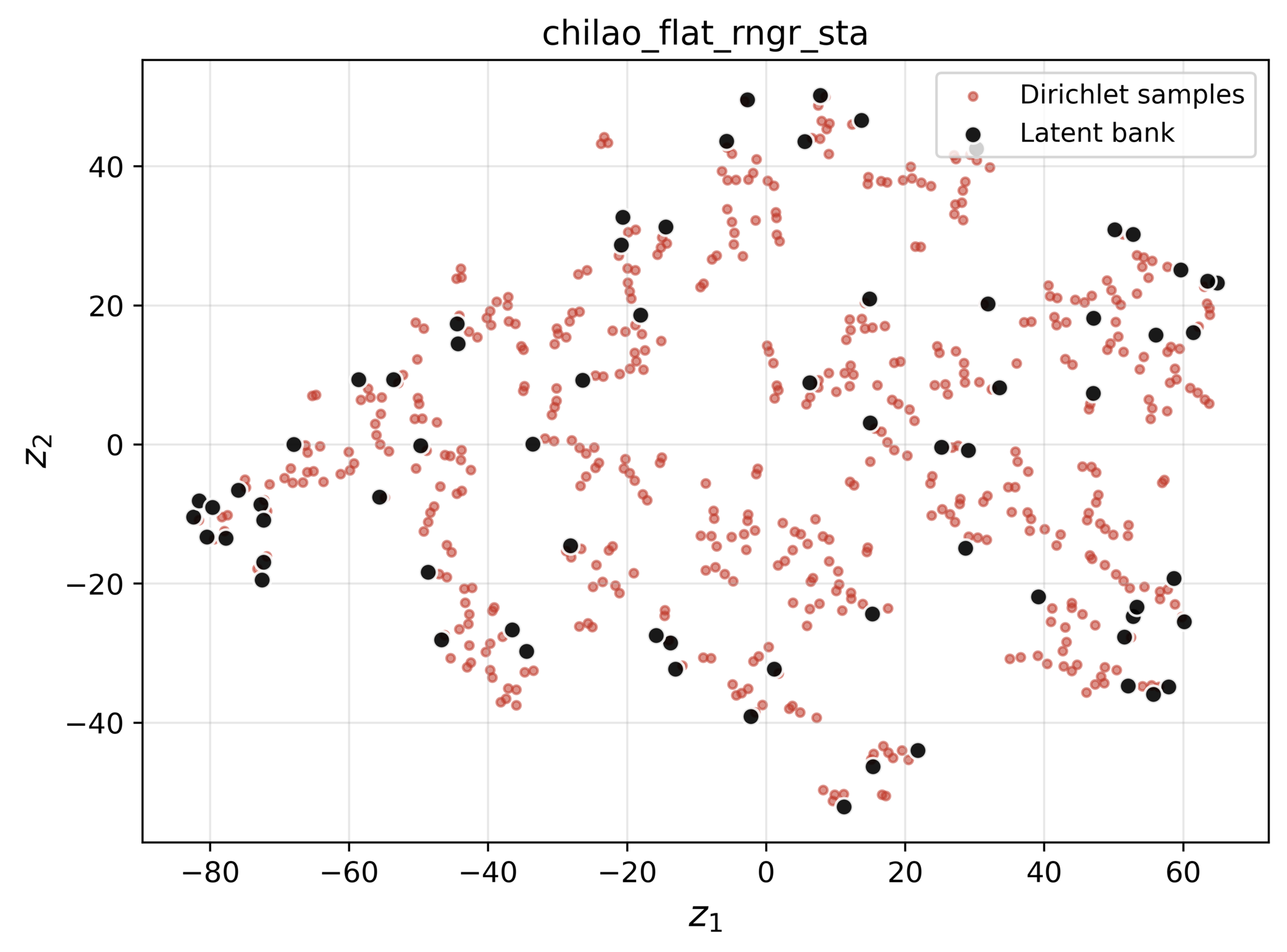}
\caption{t-SNE projection of the latent space for the Chilao station (NGA-West2). Black points represent the latent bank from real recordings; red points represent samples generated via Dirichlet-based sampling.}
\label{fig:latent_tsne_nga}
\end{figure}

Analyzing the latent space structure for cross-regional records evaluates the generalization capability of the feature extraction module. Similar to the intra-regional analysis, we project NGA-West2 data into the latent space to assess whether the model maps diverse tectonic regions into a coherent physical representation. Figure \ref{fig:latent_tsne_nga} illustrates the $t$-SNE projection for the Chilao station.

The generated samples (red points), produced via the Dirichlet-based sampling strategy, effectively cover the domain of the real records (black points). This continuous coverage indicates that the sampling mechanism captures the inherent aleatory variability within the station-specific latent bank, enabling the generation of diverse, physically bounded realizations without diverging into non-physical signal spaces.

\subsubsection{Spectral Consistency}

To verify that the generated waveforms accurately reflect local site amplification effects, we compare the generated and real HVSR distributions for the NGA-West2 sites. 

\begin{figure*}[!htbp]
\centering

\begin{subfigure}{0.38\textwidth}
    \centering
    \includegraphics[width=\linewidth]{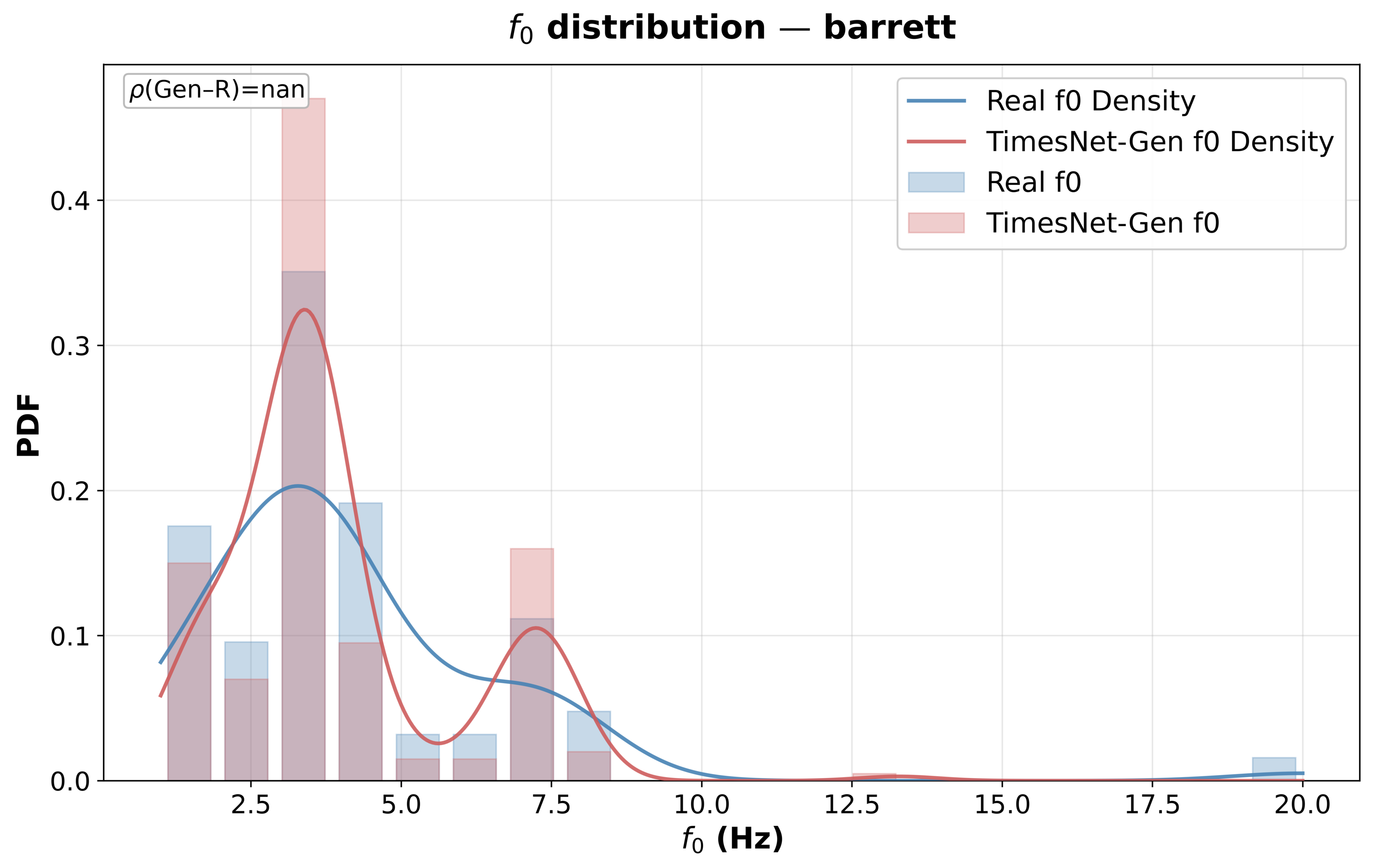}
    \caption{Barrett}
\end{subfigure}
\hspace{0.5cm}
\begin{subfigure}{0.42\textwidth}
    \centering
    \includegraphics[width=\linewidth]{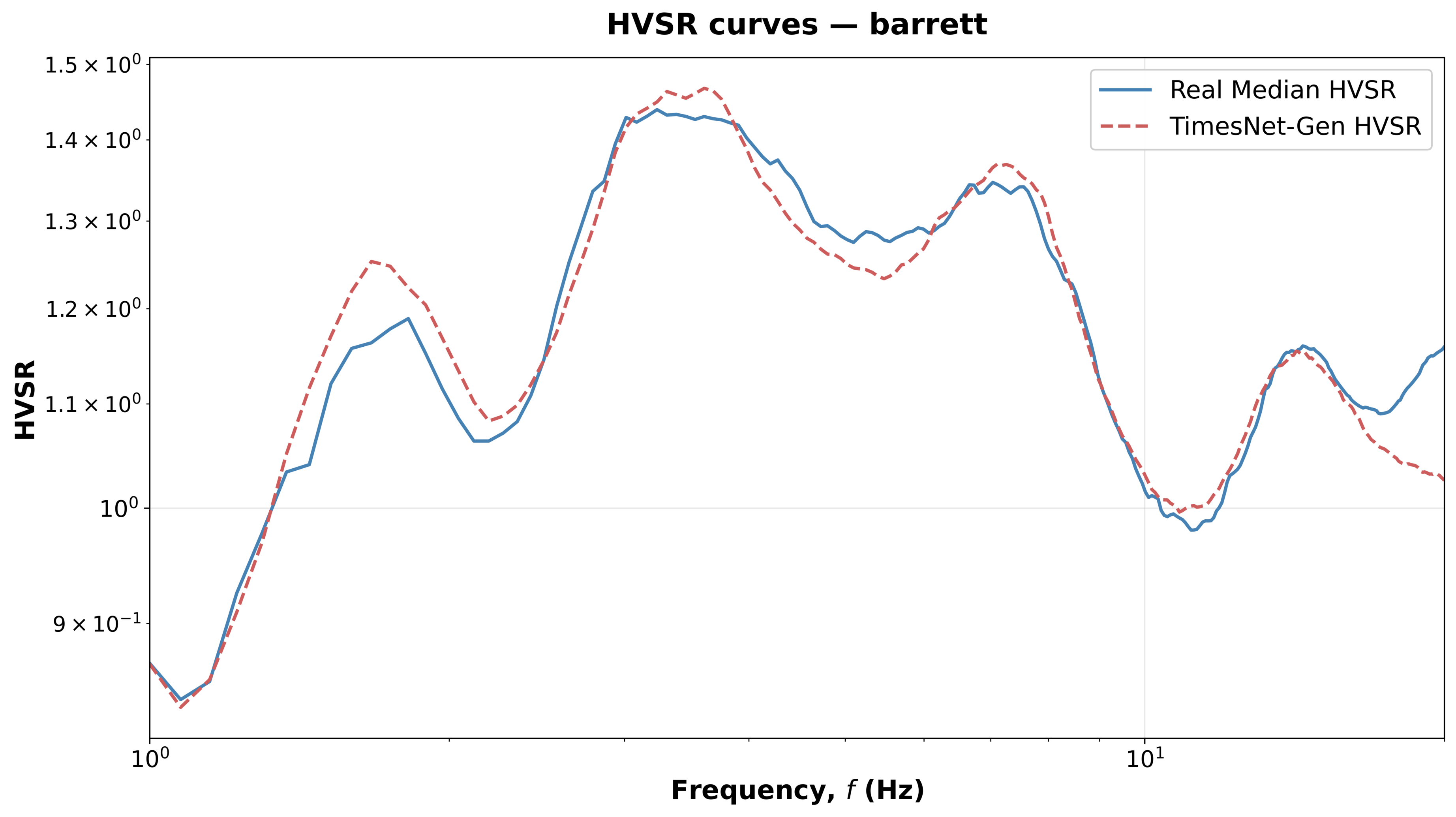}
    \caption{Barrett}
\end{subfigure}

\vspace{0.15cm} 

\begin{subfigure}{0.38\textwidth}
    \centering
    \includegraphics[width=\linewidth]{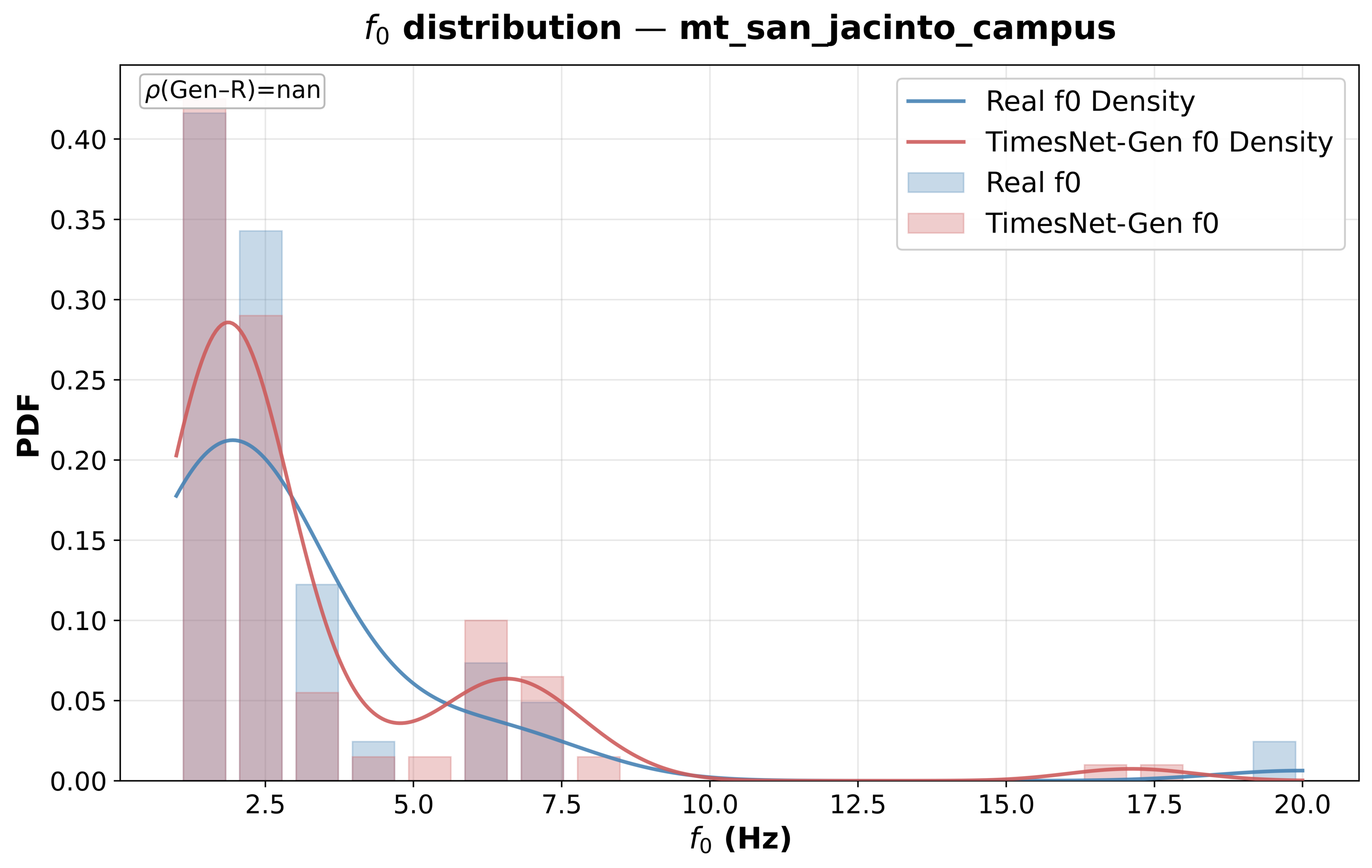}
    \caption{San Jacinto Campus}
\end{subfigure}
\hspace{0.5cm}
\begin{subfigure}{0.42\textwidth}
    \centering
    \includegraphics[width=\linewidth]{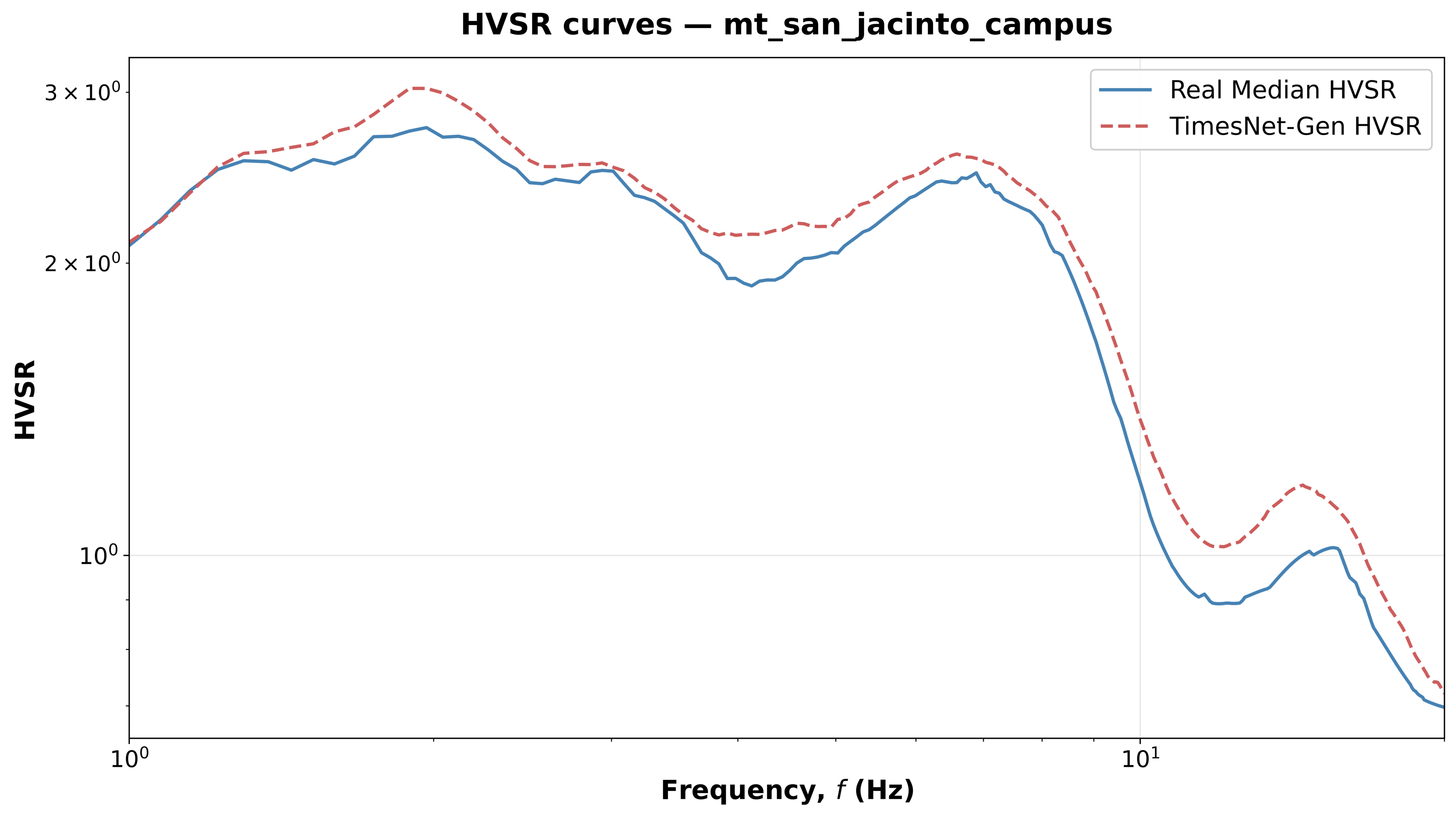}
    \caption{San Jacinto Campus}
\end{subfigure}

\vspace{0.15cm}

\begin{subfigure}{0.38\textwidth}
    \centering
    \includegraphics[width=\linewidth]{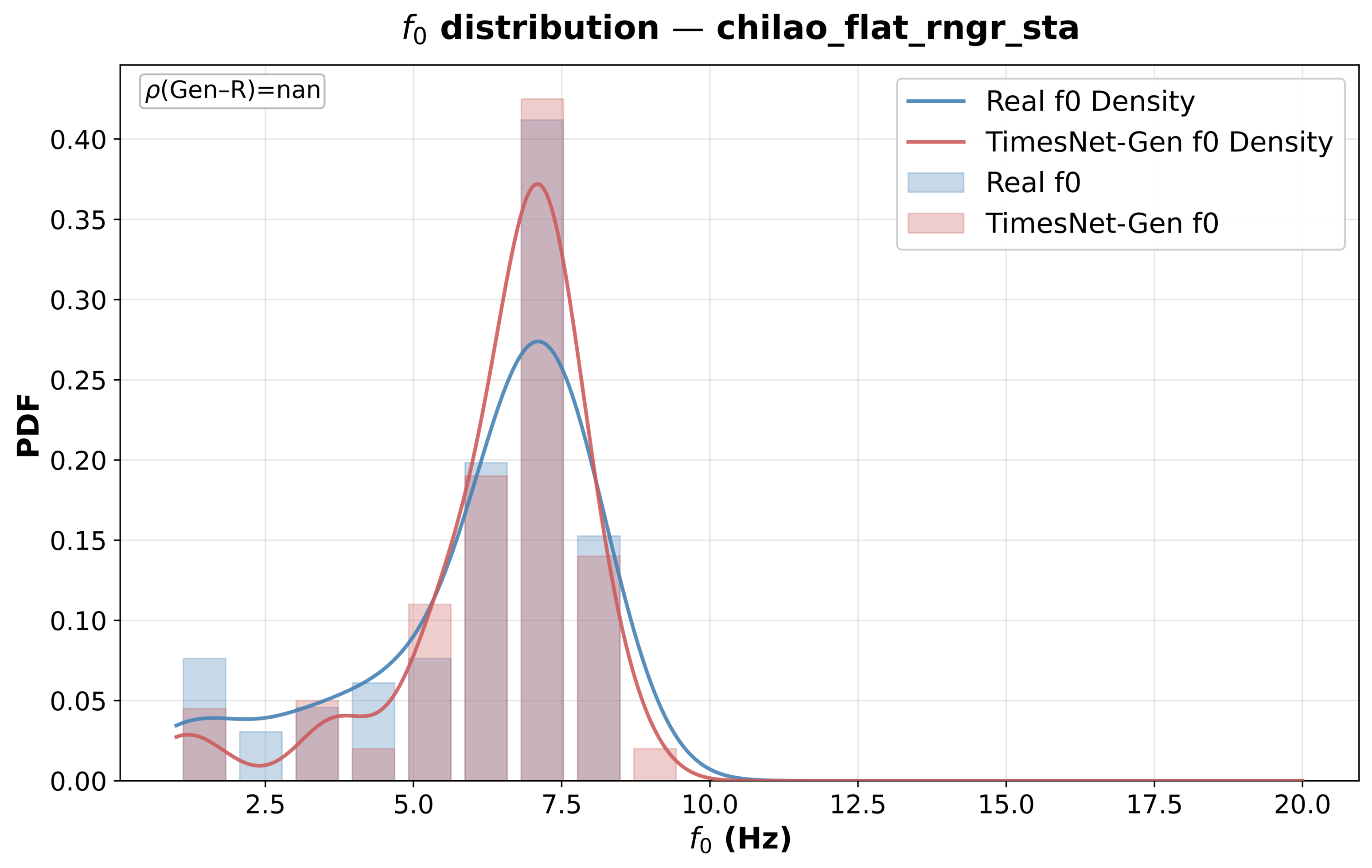}
    \caption{Chilao Flat Ranger}
\end{subfigure}
\hspace{0.5cm}
\begin{subfigure}{0.42\textwidth}
    \centering
    \includegraphics[width=\linewidth]{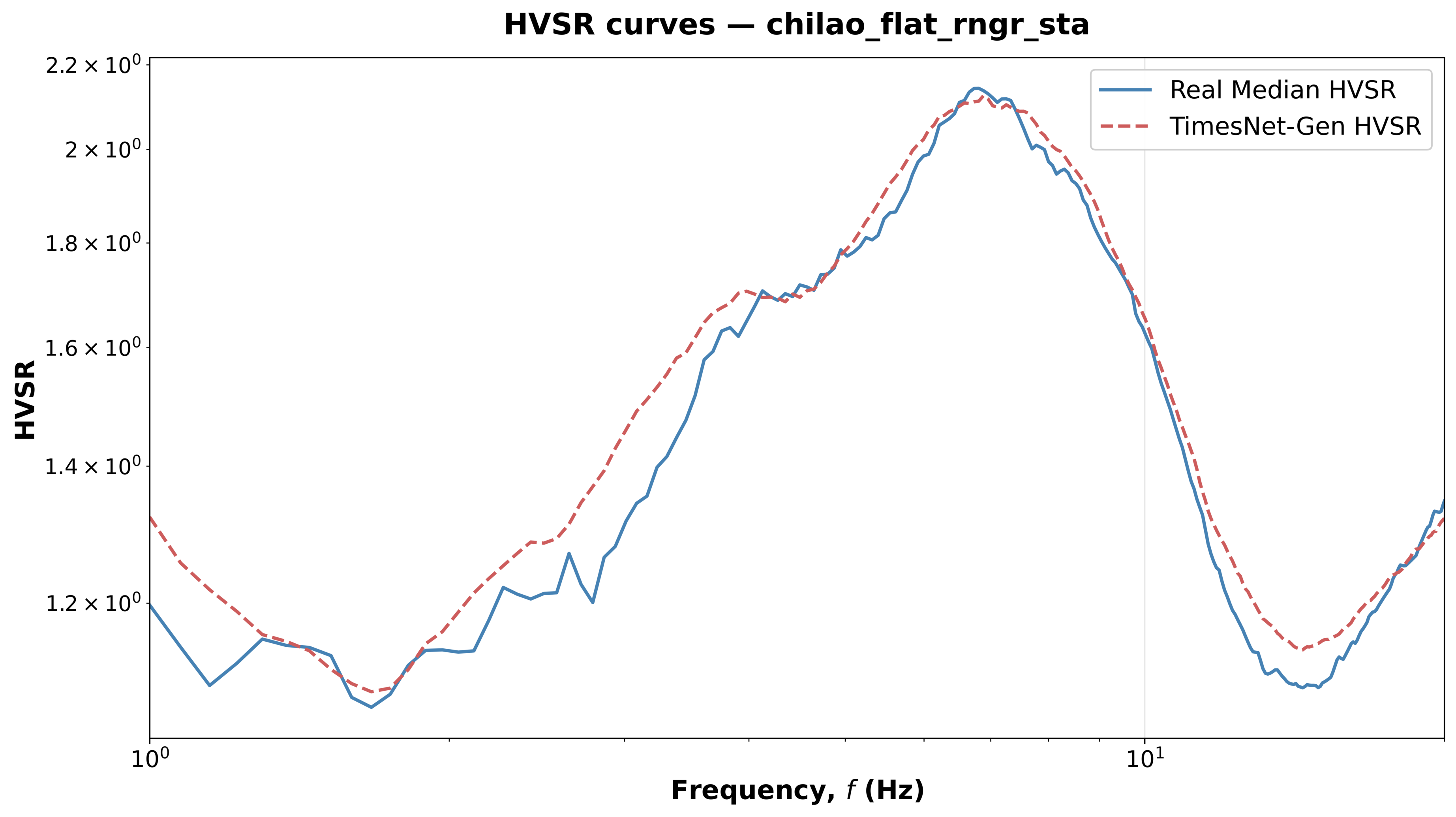}
    \caption{Chilao Flat Ranger}
\end{subfigure}

\vspace{0.15cm}

\begin{subfigure}{0.38\textwidth}
    \centering
    \includegraphics[width=\linewidth]{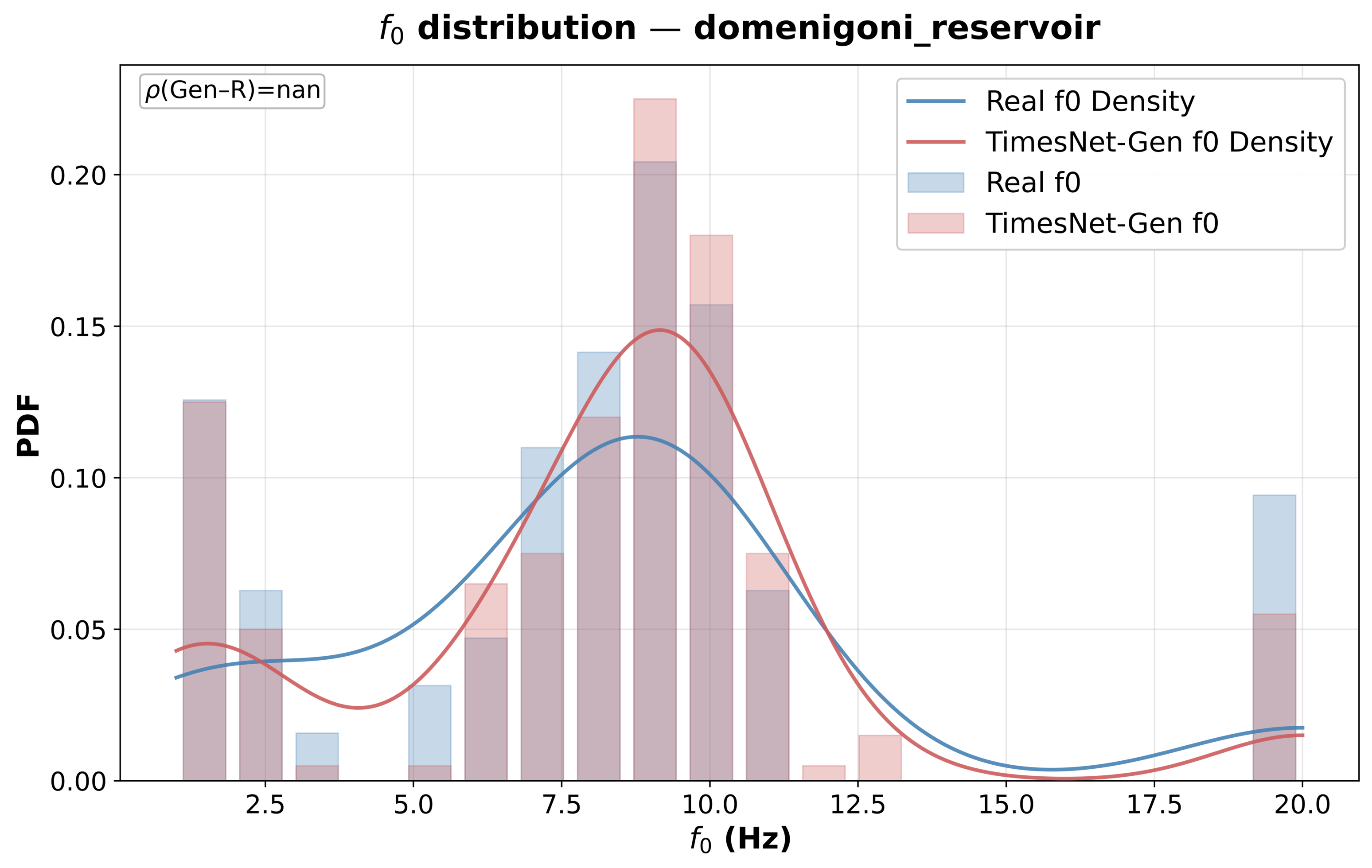}
    \caption{Domenigoni Reservoir}
\end{subfigure}
\hspace{0.5cm}
\begin{subfigure}{0.42\textwidth}
    \centering
    \includegraphics[width=\linewidth]{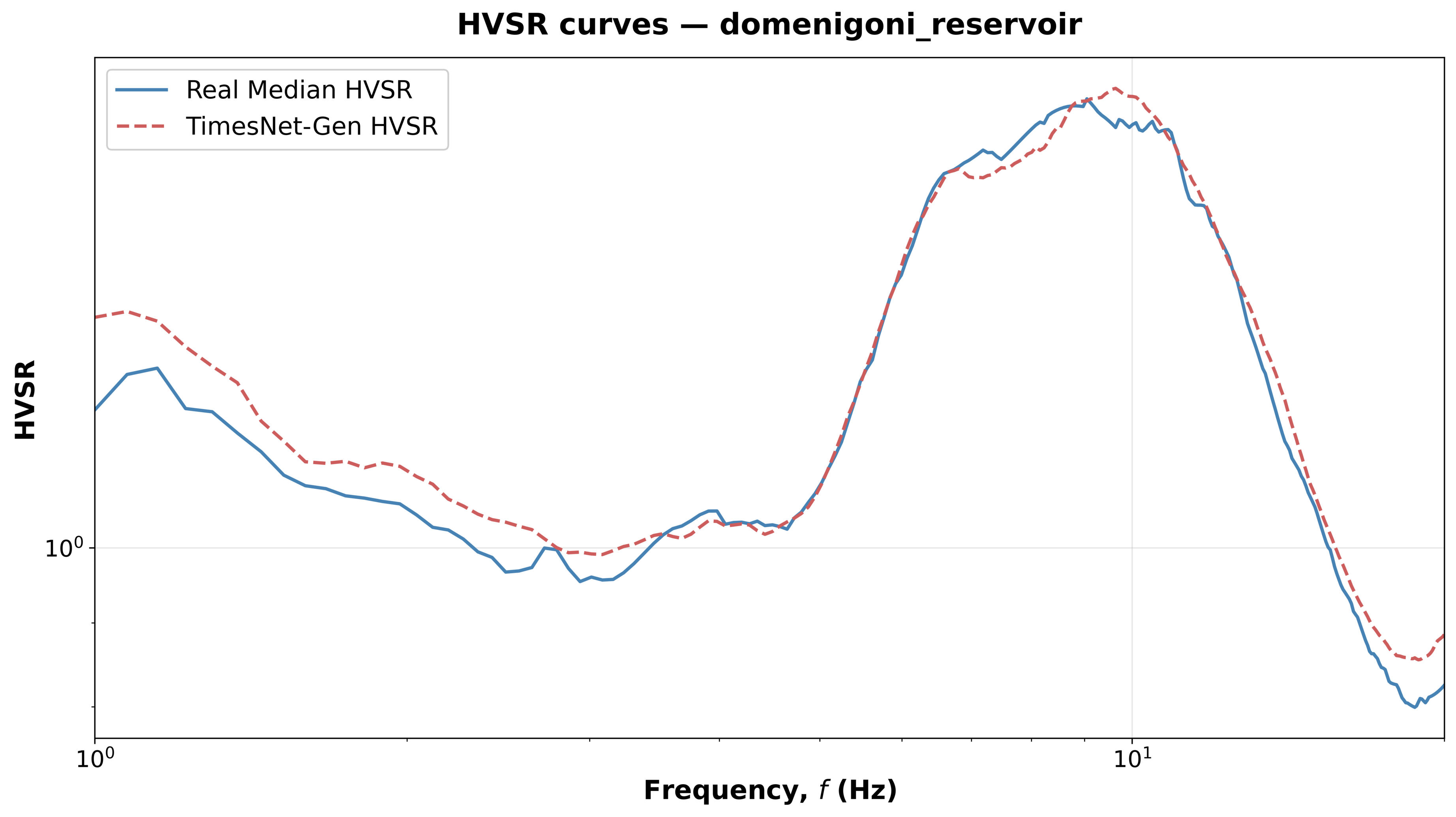}
    \caption{Domenigoni Reservoir}
\end{subfigure}

\vspace{0.15cm}

\begin{subfigure}{0.38\textwidth}
    \centering
    \includegraphics[width=\linewidth]{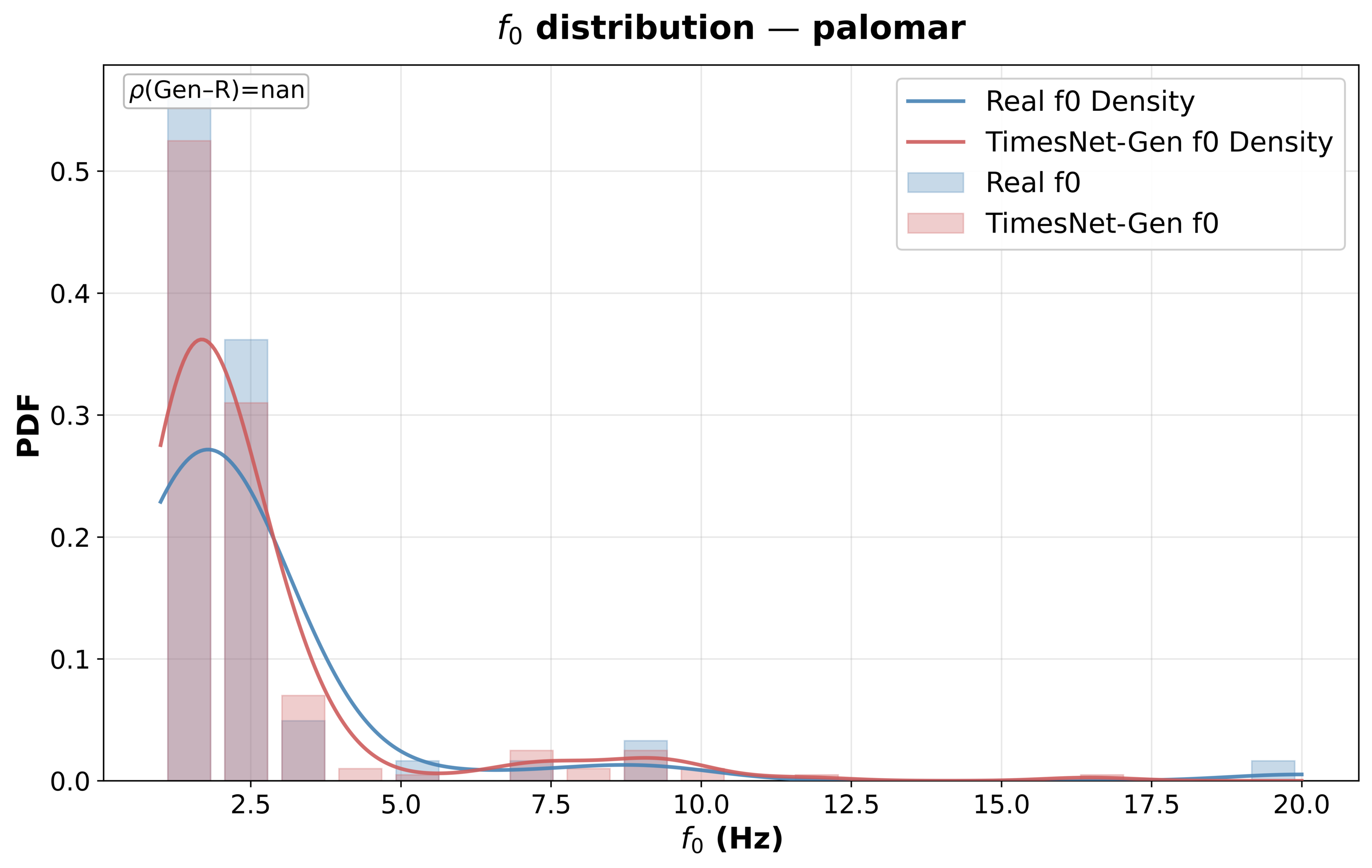}
    \caption{Palomar Observatory}
\end{subfigure}
\hspace{0.5cm}
\begin{subfigure}{0.42\textwidth}
    \centering
    \includegraphics[width=\linewidth]{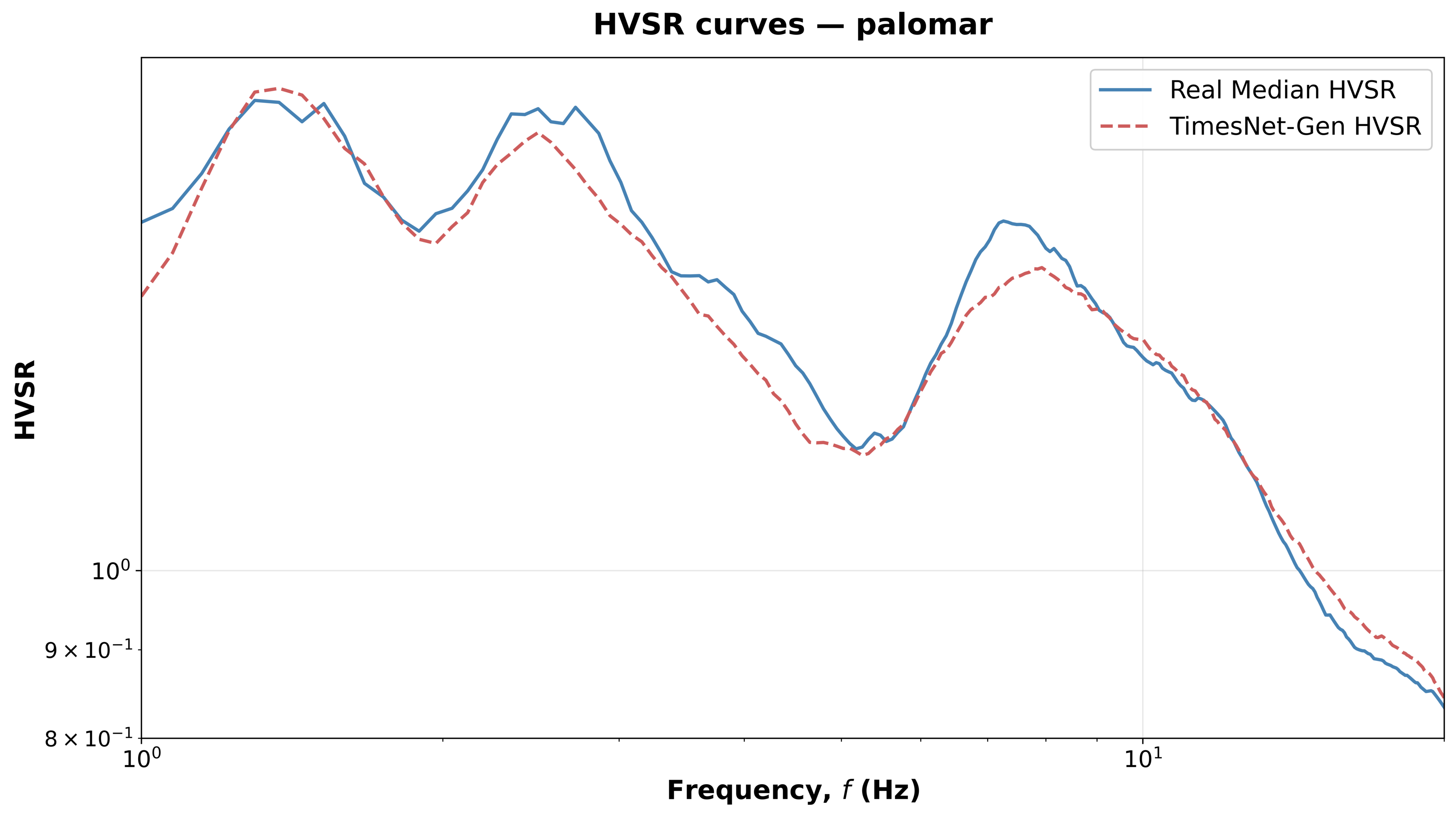}
    \caption{Palomar Observatory}
\end{subfigure}

\caption{Comparison of \texorpdfstring{$f_0$}{f0} distributions (left) and average HVSR curves (right) for the selected NGA-West2 stations.}
\label{fig:nga_combined_spectral}
\end{figure*}

Figure \ref{fig:nga_combined_spectral} demonstrates that TimesNet-Gen reproduces the overall shape and predominant resonance peaks of the HVSR curves across the selected stations. The alignment between the empirical and generated spectral responses indicates that the features learned from the AFAD database effectively reflect fundamental site response characteristics. Furthermore, generating waveforms that qualitatively match the empirical spectral distribution demonstrates that the model captures the natural variability essential for probabilistic seismic hazard applications.

\begin{figure}[htbp]
\centering

\begin{subfigure}{0.73\linewidth}
    \centering
    \includegraphics[width=\linewidth]{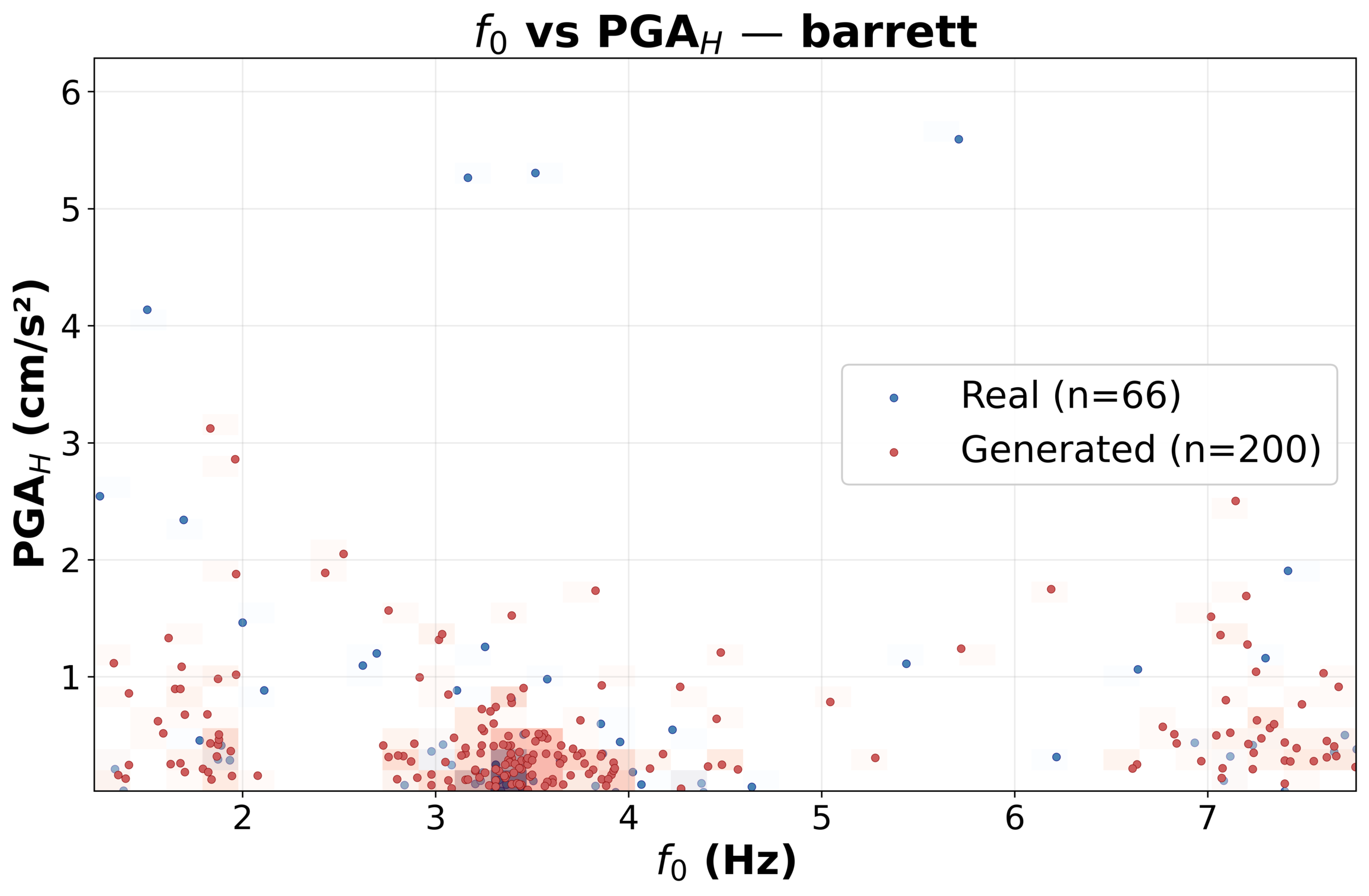}
    \caption{Barrett}
\end{subfigure}

\vspace{0.4cm} 

\begin{subfigure}{0.73\linewidth}
    \centering
    \includegraphics[width=\linewidth]{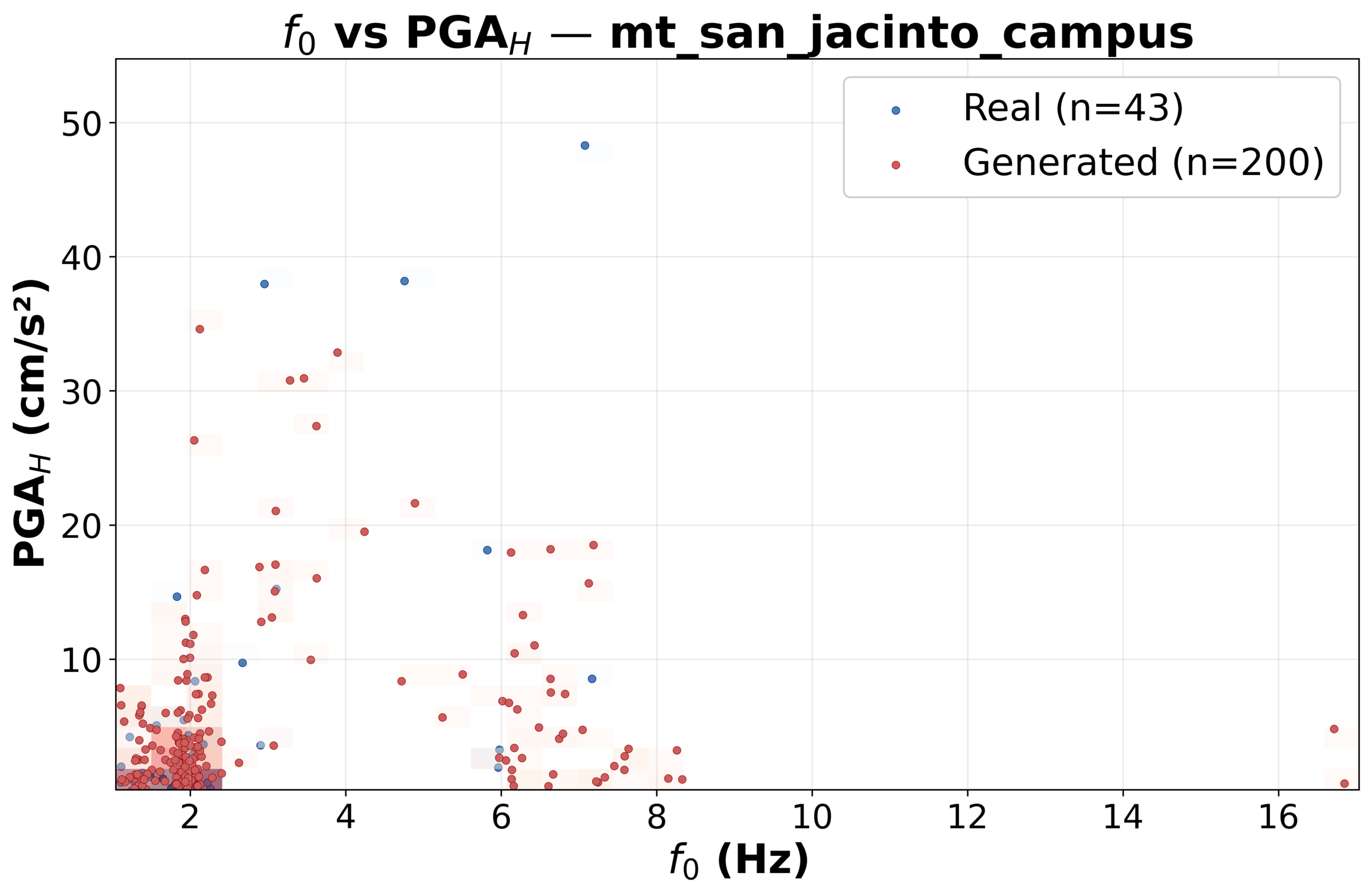}
    \caption{San Jacinto Campus}
\end{subfigure}

\vspace{0.4cm}

\begin{subfigure}{0.73\linewidth}
    \centering
    \includegraphics[width=\linewidth]{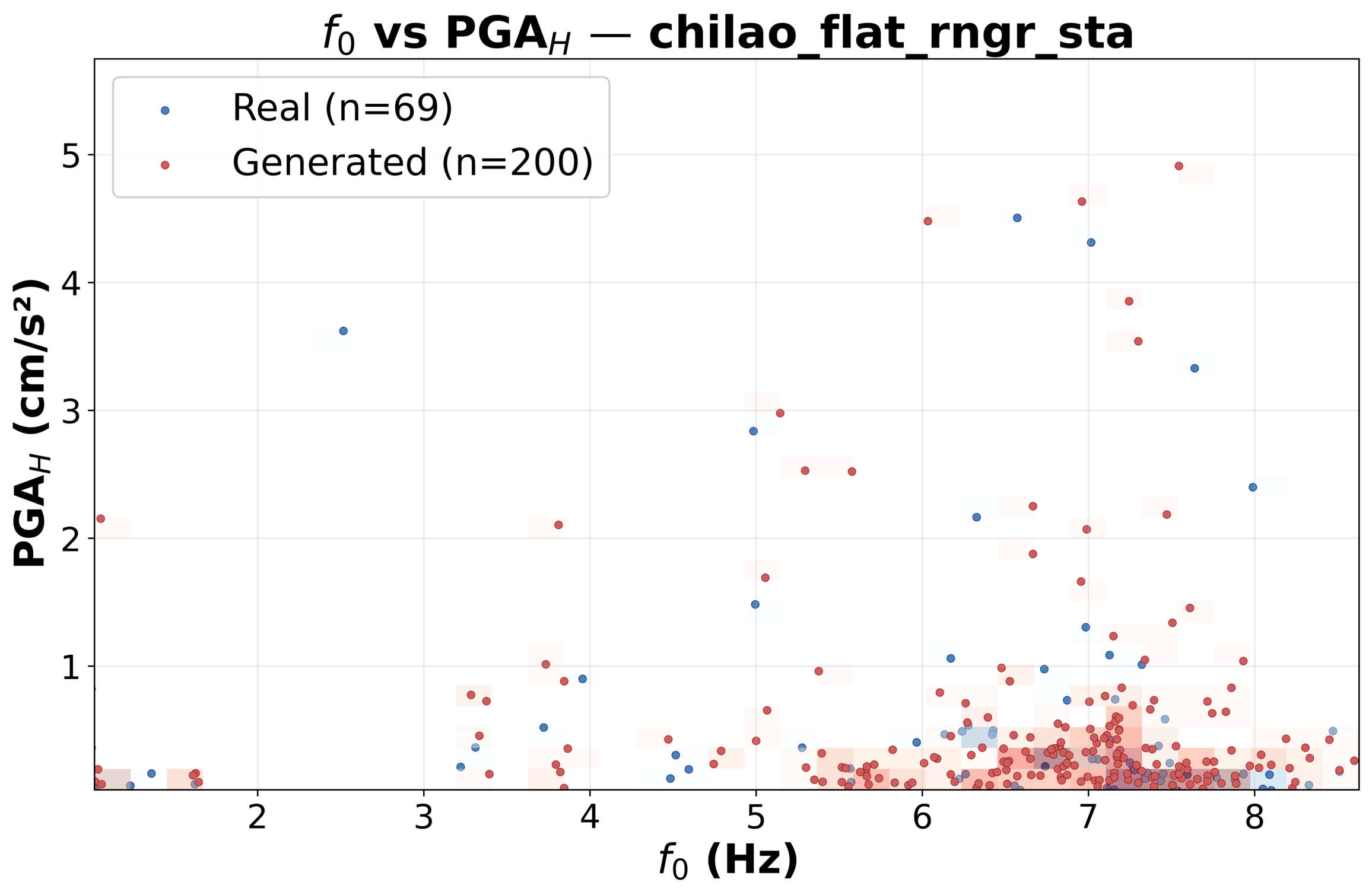}
    \caption{Chilao Flat Ranger}
\end{subfigure}

\vspace{0.4cm}

\begin{subfigure}{0.73\linewidth}
    \centering
    \includegraphics[width=\linewidth]{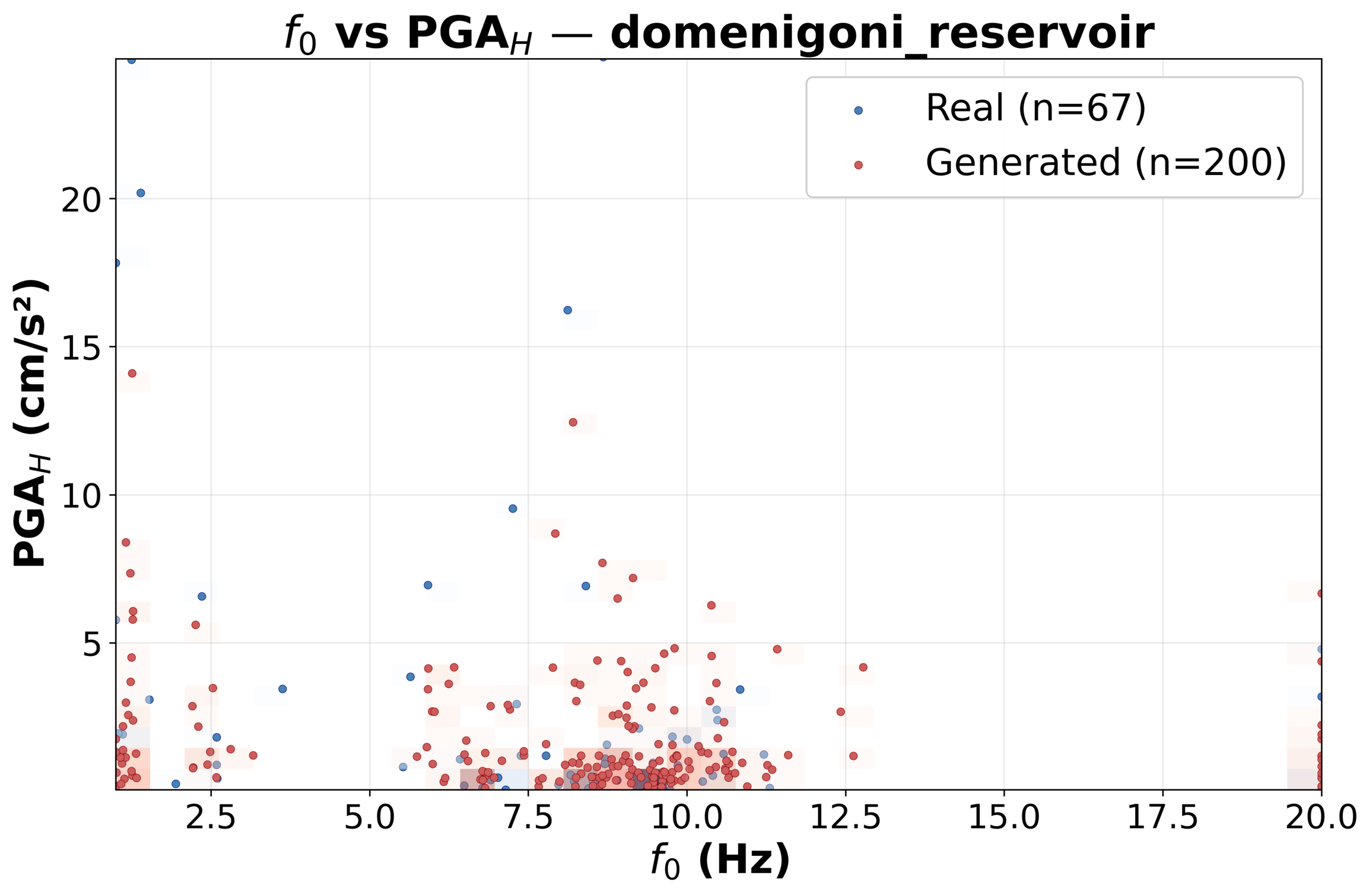}
    \caption{Domenigoni Reservoir}
\end{subfigure}

\vspace{0.4cm}

\begin{subfigure}{0.73\linewidth}
    \centering
    \includegraphics[width=\linewidth]{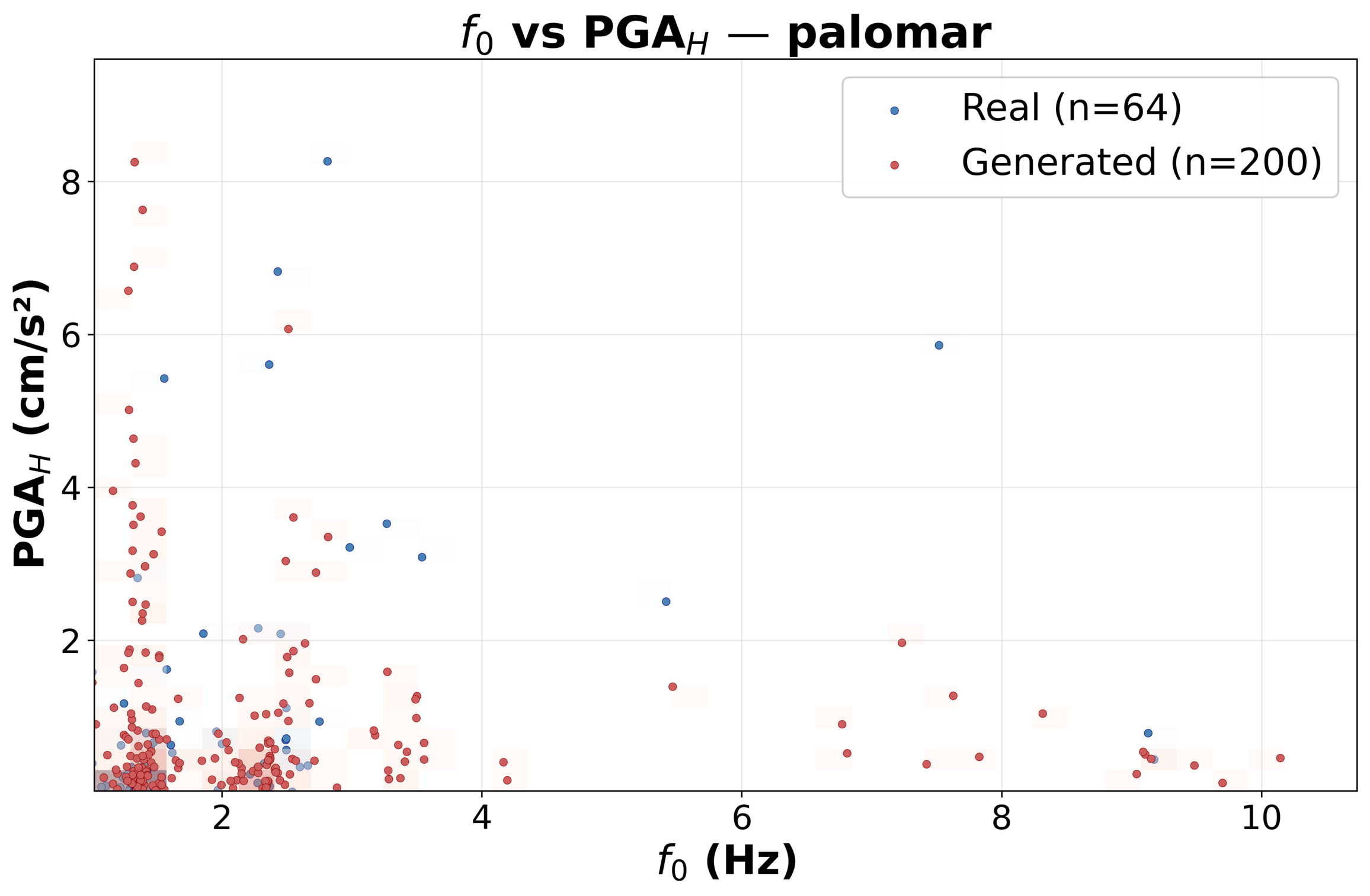}
    \caption{Palomar Observatory}
\end{subfigure}

\caption{Bivariate distributions of \texorpdfstring{$f_0$}{f0} and horizontal PGA for selected NGA-West2 stations.}
\label{fig:nga_f0_pga_grid}
\end{figure}

\subsubsection{Joint Distribution of Site Resonance and Peak Ground Acceleration}
While HVSR analysis validates the frequency content, it does not constrain absolute energy levels in the time domain. Beyond spectral consistency, generating physically viable ground motions requires preserving the inherent dependency between amplitude and frequency content. To evaluate this physical realism, we analyze the relationship between $f_0$ and horizontal Peak Ground Acceleration (PGA). In real seismic events, spectral characteristics and peak ground shaking are strongly coupled through source characteristics, path attenuation, and site-specific amplification. 

Figure \ref{fig:nga_f0_pga_grid} illustrates the bivariate joint distribution of $f_0$ and horizontal PGA for the selected NGA-West2 stations. The results demonstrate a significant qualitative overlap between the synthesized and empirical records across these geographically distinct sites. This distributional alignment confirms that TimesNet-Gen successfully preserves the essential physical coupling between resonance characteristics and peak intensity, ensuring the generated ground motions remain within physically plausible bounds for the target environments.

\subsubsection{Cross-Regional Interstation Discrimination}
\begin{figure}[!htbp]
    \centering
    \includegraphics[width=0.8\columnwidth,trim=0 75 300 90,clip]{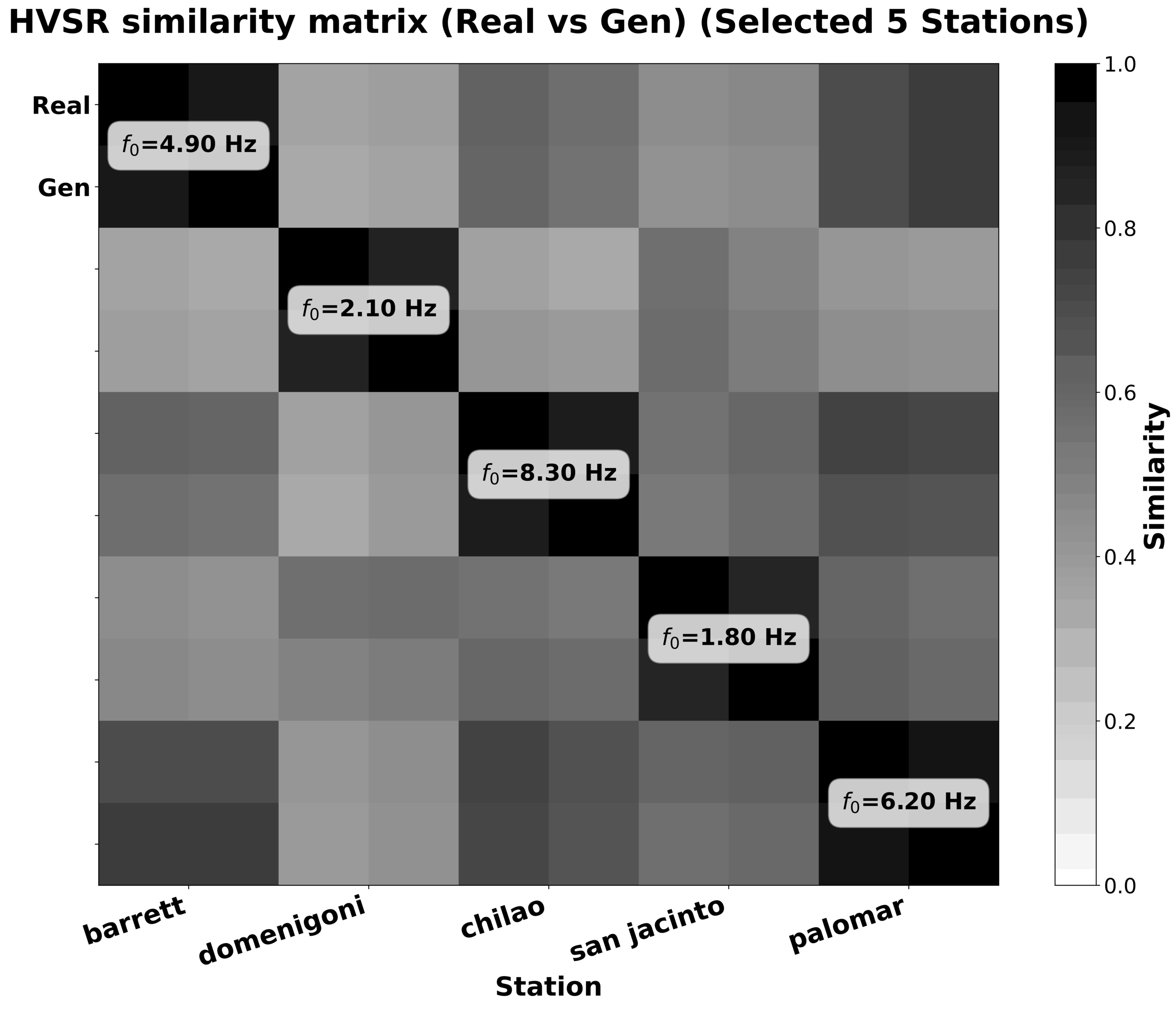}
    \caption{TimesNet-Gen (alignment score: 0.82) for NGA-West2 stations.}
    \label{fig:nga_jsd_matrix}
\end{figure}

To evaluate the model's physical consistency and discrimination capability without regional fine-tuning, we extend the $f_{0}$ distribution confusion matrix analysis to the NGA-West2 dataset. Figure \ref{fig:nga_jsd_matrix} presents the JSD similarity scores between the empirical and generated distributions for the Southern California stations. Similar to the intra-regional evaluation, several NGA-West2 sites exhibit comparable fundamental frequencies. The resulting matrix demonstrates that TimesNet-Gen successfully distinguishes these geographically distinct stations, achieving an alignment score of 0.82 against an ideal identity-block matrix. This strong cross-regional separation confirms that the latent representation does not merely overfit to the training region, but actively captures generalized feature representations necessary to isolate individual station characteristics, using only the empirical bank of the target region.

\section{Conclusions}
In this study, we introduced TimesNet-Gen, a generative framework designed to synthesize site-specific strong ground motion records directly in the time domain through a station-restricted latent resampling strategy. By avoiding the restrictive parametric assumptions of traditional stochastic modeling, the proposed approach learns the complex, nonlinear characteristics of seismic sites directly from empirical data distributions.

The primary findings and contributions of this work are summarized as follows. First, we implemented a Dirichlet-based latent space sampling mechanism that couples latent embeddings with their instance-level scaling parameters. This strategy effectively guides the generative process, mitigating the risk of producing physically inconsistent amplitude scaling while preserving the aleatory variability reflected in the target station's empirical latent bank. Second, we evaluated parameter-level transferability by applying a frozen, AFAD-trained encoder to the geographically and tectonically distinct NGA-West2 dataset. The ability to encode NGA-West2 records into station-specific latent banks in Southern California without any regional fine-tuning or weight updates indicates that TimesNet-Gen learns broadly transferable waveform representations rather than merely memorizing the AFAD training distribution. Third, using log-HVSR distribution comparisons and joint $f_0$--PGA analysis, we demonstrated that the synthesized records remain consistent with the empirical coupling between site resonance and peak intensity under the reported metrics. Furthermore, empirical comparisons show that TimesNet-Gen outperforms the spectrogram-based conditional VAE baseline in capturing sharp resonance peaks and non-stationary temporal patterns.

Beyond these methodological contributions, TimesNet-Gen is well suited to support single-station machine learning pipelines and the generation of probabilistic catalogs under intensity measure and HVSR checks. The framework is particularly advantageous for augmenting data at sparse stations and providing a pre-design exploratory synthesis pool. By capturing site-specific variations at the distribution level, this data-driven approach can effectively serve as a training data enrichment tool for downstream deep learning models.

Despite these promising results, several limitations should be acknowledged to contextualize the framework's current capabilities. The formulation requires an existing latent bank constructed from empirical recordings at the target station, limiting its direct applicability to uninstrumented sites. Moreover, the present model does not incorporate explicit conditioning on earthquake magnitude ($M$) or rupture distance ($R$), which restricts it to probabilistic sampling rather than scenario-controlled generation. However, by successfully establishing a stable, physics-consistent latent manifold that captures site-specific waveform statistics, we have laid the foundational groundwork for these extensions. We view this work as a critical first step toward scenario-conditioned generative workflows.

Building directly upon this architecture, future work will focus on integrating explicit physical constraints, such as $M$, $R$, and $V_{s30}$, into the sampling and decoding pipeline. Ultimately, this data-driven approach does not seek to replace physics-based simulations for regulated seismic hazard analysis, but rather aims to provide a complementary data layer. Once explicit conditioning is achieved, the framework's outputs can be directly benchmarked against established physics-based simulations such as those on the SCEC Broadband Platform under shared scenario definitions and a broader suite of intensity measures. In the longer term, integrating the high-frequency site variability learned directly from empirical recordings holds the potential to supplement the low-frequency reliability of physics-based models, paving the way for advanced hybrid broadband ground-motion synthesis.
\section{Acknowledgements}
This study was supported by the Middle East Technical University (METU) Scientific Research Projects (BAP) under grant number ADEP-704-2024-11482.

\bibliography{refs}

\end{document}